\documentclass{article}

\usepackage{microtype}
\usepackage{graphicx}
\usepackage{subfigure}
\usepackage{booktabs} 
\usepackage{multirow}

\usepackage{hyperref}
\usepackage{booktabs}
\usepackage{pifont}
\newcommand{\cmark}{\ding{51}}
\newcommand{\xmark}{\ding{55}}

\usepackage[accepted]{icml2026}

\usepackage{amsmath}
\usepackage{amssymb}
\usepackage{mathtools}
\usepackage{amsthm}

\usepackage[capitalize,noabbrev]{cleveref}

\theoremstyle{plain}
\newtheorem{theorem}{Theorem}[section]

\newtheorem{lemma}[theorem]{Lemma}

\theoremstyle{definition}
\newtheorem{definition}[theorem]{Definition}

\theoremstyle{remark}
\newtheorem{remark}[theorem]{Remark}

\usepackage[textsize=tiny]{todonotes}
\usepackage{subcaption}

\icmltitlerunning{Progressively Deformable 2D Gaussian Splatting for Video Representation at Arbitrary Resolutions}

\begin{document}

\twocolumn[
\icmltitle{Progressively Deformable 2D Gaussian Splatting for Video Representation at Arbitrary Resolutions}



\icmlsetsymbol{equal}{*}

\begin{icmlauthorlist}
\icmlauthor{Mufan Liu}{equal,sjtu}
\icmlauthor{Qi Yang}{equal,umkc}
\icmlauthor{Miaoran Zhao}{sjtu}
\icmlauthor{He Huang}{sjtu}
\icmlauthor{Le Yang}{ucan}
\icmlauthor{Zhu Li}{umkc}
\icmlauthor{Yiling Xu}{sjtu}
\end{icmlauthorlist}

\icmlaffiliation{sjtu}{Shanghai Jiao Tong University, Shanghai, China}
\icmlaffiliation{umkc}{University of Missouri--Kansas City, Kansas City, MO, USA}
\icmlaffiliation{ucan}{University of Canterbury, Christchurch, New Zealand}

\icmlcorrespondingauthor{Yiling Xu}{yl.xu@sjtu.edu.cn}


\icmlkeywords{Machine Learning, ICML}

\vskip 0.3in
]




\printAffiliationsAndNotice{\icmlEqualContribution}


\begin{abstract}
Implicit neural representations (INRs) enable fast video compression and effective video processing, but a single model rarely offers scalable decoding across rates and resolutions. In practice, multi-resolution typically relies on retraining or multi-branch designs, and structured pruning failed to provide a permutation-invariant progressive transmission order. Motivated by the explicit structure and efficiency of Gaussian splatting, we propose D2GV-AR, a deformable 2D Gaussian video representation that enables \emph{arbitrary-scale} rendering and \emph{any-ratio} progressive coding within a single model. We partition each video into fixed-length Groups of Pictures and represent each group with a canonical set of 2D Gaussian primitives, whose temporal evolution is modeled by a neural ordinary differential equation.
During training and rendering, we apply scale-aware grouping according to Nyquist sampling theorem to form a nested hierarchy across resolutions. Once trained, primitives can be pruned via a D-optimal subset objective to enable any-ratio progressive coding. Extensive experiments show that D2GV-AR renders at over 250 FPS while matching or surpassing recent INR baselines, enabling multiscale continuous rate--quality adaptation.
\end{abstract}

\begin{figure}[t]
    \centering
    \includegraphics[width=\linewidth]{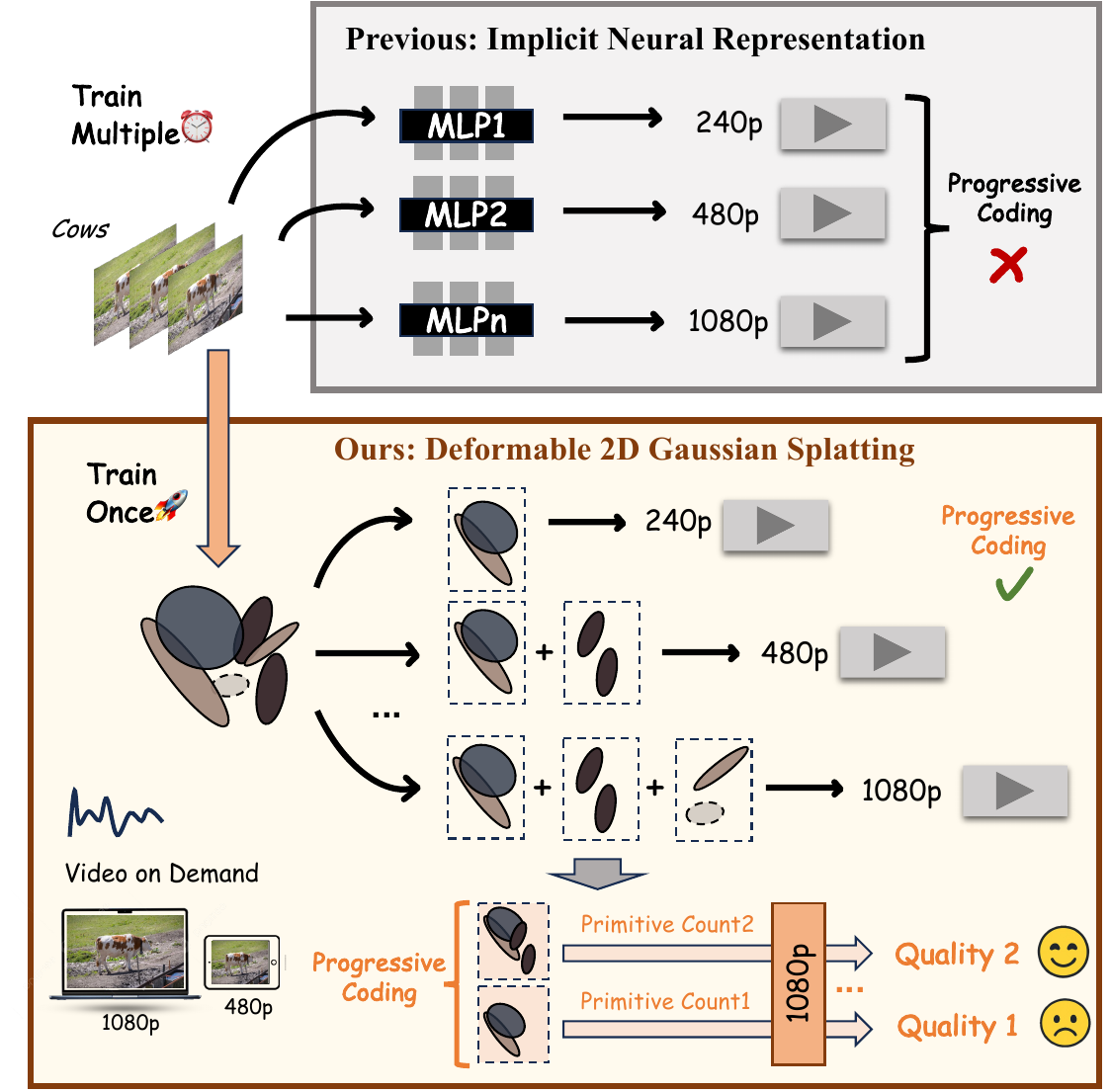}
        \vspace{-0.2cm}
    \caption{Design motivation. INR-based media usually depends on discrete but finite resolution branches, whereas our model is trained once and supports arbitrary resolutions with progressive coding for continuous rate adaptation.}
    \vspace{-0.6cm}
    \label{fig:motiv}
\end{figure}

\section{Introduction}
A key challenge in video-on-demand (VoD) delivery is \emph{heterogeneity}: end devices differ in display resolution and computing capacity, while rapid bandwidth fluctuations calls for timely bitrate adaptation. This motivates the use of multi-resolution decoding and on-the-fly rate control from a single video representation. However, most mainstream codecs (e.g., H.264/AVC \cite{wiegand2003overview} and H.265/HEVC \cite{pastuszak2015algorithm}) are optimized for a single operating point; supporting diverse devices and dynamic network conditions thus often requires storing and transmitting multiple versions of the same content.

Scalable video coding (SVC) \cite{schwarz2007svc} is an early but widely adopted solution for multi-resolution rate adaptation. It extends the canonical codecs with layered bitstreams, allowing a receiver to decode a subset of layers that matches its hardware resources and available bandwidth. More recently, implicit neural representations (INRs) \cite{chen2021nerv,chen2023hnerv} have been studied as a VoD-friendly media format. They parameterize a video segment as a continuous function modeled by a neural network, enabling fast decoding with a simple pipeline. INR-based scalability follows two types of designs: a single cascaded network with intermediate prediction heads for multiple resolutions \cite{chen2021nerv}, or resolution-specific branches with shared sublayer computation \cite{wei2024snerv}. They both rely on fixed architectures and operate at a discrete set of operating points, which limits their flexibility and hinders more graceful rate adaptation.


Recently, Gaussian splatting (GS) has represented 3D visual content with anisotropic primitives, as well as effective rendering via CUDA rasterization. The same property applies to 2D as well: optimizing “flattened” Gaussians directly in the image plane yields high-quality reconstructions with fast decoding. Unlike pruning neural networks, which is architecture-dependent and often requires additional bookkeeping to track removed parameters, GS is naturally scalable: each primitive has an explicit pixel footprint, enabling resolution-aware subset selection, and progressive refinement is achievable by simply retaining more primitives. 

Despite the inherent advantages of GS, a streamable GS-based video representation needs to address the following two challenges, which are also the contributions of this paper. Despite the practical advantages of Gaussian splatting, a \emph{streamable} GS-based video representation must address two challenges.
\textbf{First, compactness:} videos reuse content over time; per-frame Gaussians without sharing waste capacity on near-duplicate structures. \textbf{Second, fixed rate adaptation:} existing scalable designs, either layered bitstreams or branch-based implicit neural representations, expose only a few fixed bitrate--resolution operating points (See Fig. \ref{fig:motiv}).

\begin{figure}[t]
    \centering
    \includegraphics[width=\linewidth]{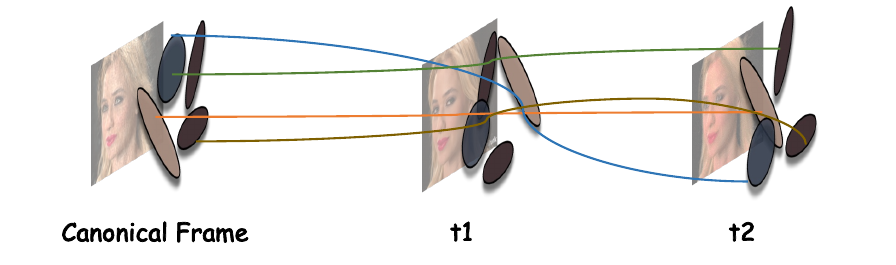}
    \vspace{-0.4cm}
    \caption{A toy example of deformable 2D Gaussian field.}
        \vspace{-0.8cm}
    \label{toydeform}
\end{figure}

In this paper, we propose the progressively \textbf{D}eformable \textbf{2}D \textbf{G}aussian splatting for \textbf{V}ideo representation at \textbf{A}rbitrary \textbf{R}esolutions (D2GV-AR), a streamable representation that supports arbitrary-scale rendering and progressive coding for on-the-fly rate adaptation. \textbf{For Challenge 1,} we split each video into fixed-length Groups of Pictures (GoP) and learn an independent representation per GoP, which stabilizes optimization and keeps complexity manageable as the sequence length grows. Within a GoP, D2GV-AR learns a canonical set of 2D Gaussian primitives and renders each frame by deforming this canonical set over time (Fig.~\ref{toydeform}); the deformation is parameterized by a neural ordinary differential equation (neural ODE) to model the continuous-time evolution and encourage temporal coherence. \textbf{For Challenge 2,} we introduce \emph{scale-aware grouping} for arbitrary-scale decoding and \emph{D-optimal pruning} for progressive coding. Guided by Nyquist sampling criterion, we set a per-scale anti-aliasing cutoff and keep only Gaussians whose spectra lie below it, yielding a nested subset hierarchy used for both training and rendering; this multiscale supervision also strengthens progressive coding. After training, we enable progressive coding by pruning low-impact primitives. Prior work \cite{fan2023lightgaussian} shows that primitives covering more pixels tend to carry more signal. In 2D, Gaussians provide no occlusion, so usefulness depends on their non-overlapped footprint. We therefore pose pruning as D-optimal subset selection over the induced Gaussian kernels and use a surrogate ranker that balances footprint area against a coherence penalty. Experiments show that D2GV-AR matches INR baselines at both single and multiple resolutions, while CUDA-based differentiable rasterization substantially accelerates decoding; D2GV-AR generalizes to downstream tasks such as video interpolation. \\The contributions of our work are:
\begin{itemize}
\item We propose D2GV-AR, a streamable deformable 2D Gaussian video representation trained at the GoP level;
\item We enable arbitrary-scale decoding via scale-aware grouping, and continuous rate adaptation via progressive D-optimal pruning with a lightweight per-primitive score;
\item Experiments on Bunny, DAVIS, and UVG show competitive quality, substantially faster decoding, and improved progressive rate adaptation over INR baselines.
\end{itemize}

\section{Related Work}
\label{sec:related}
\textbf{Implicit Neural Representations: }INRs model multimedia signals as continuous coordinate-to-value functions and have been explored for view synthesis and image/video representation. Early video INRs regress pixels directly (e.g., SIREN~\cite{sitzmann2020implicit}), while later works predict full frames from frame indices (NeRV~\cite{chen2021nerv}) and improve efficiency/quality via factorized spatiotemporal representations and content-adaptive embeddings~\cite{li2022nerv,chen2022cnerv,chen2023hnerv}. More recent variants introduce feature grids or disentangled codes to strengthen latent representations and capture dynamics ~\cite{lee2023ffnerv,kwan2024hinerv,yan2024ds}. Despite steady progress, INR-based video models remain computation-heavy to train and decode, and their implicit parameterization offers limited interpretability and controllability, which makes scalable video representation challenging.\\
\textbf{Gaussian Splatting:} GS has recently emerged as a powerful approach for 3D view synthesis \cite{3dgs}, leveraging explicit 3D Gaussian representations and differentiable tile-based rasterization to enable real-time rendering and enhanced editability \cite{sun2024splatter}. Its effectiveness has led to widespread adoption in dynamic scene modeling \cite{yang2024deformable, wu20244d, li2024spacetime}, AI-generated content \cite{tang2023dreamgaussian, yi2023gaussiandreamer}, autonomous driving \cite{zhou2024drivinggaussian, zielonka2023drivable} and wireless communication \cite{wen2024wrf, zhang2024rf}. While primarily designed for 3D applications, GS has also been explored in the 2D domain. GaussianImage \cite{zhang2024gaussianimage} utilizes 2D Gaussian primitives for fast image representation, while \cite{hu2024gaussiansr} replaces convolution kernels with 2D Gaussians to enhance arbitrary-scale image super-resolution.\\
\textbf{Scalable Video Coding: }
Scalable video coding organizes a bitstream into a base layer and enhancement layers for adaptive decoding, as standardized by the scalable extensions of H.264/AVC (SVC~\cite{schwarz2007svc}) and H.265/HEVC (SHVC~\cite{schwarz2007svc}).
In real-time streaming, VP9 SVC~\cite{vp9bitstream2016} is widely adopted (e.g., WebRTC) to expose a small set of temporal/spatial layers for bandwidth adaptation. Scalability has also been explored in implicit neural representations by structuring the decoder itself, e.g., NeRV-style intermediate heads~\cite{NeRV_GitHub_2021}, base-plus-enhancement transmission in QS-NeRV~\cite{wu2024qs}, and explicit enhancement branches in SNeRV~\cite{wei2024snerv}.
In contrast, we leverage explicit Gaussian rasterization to support arbitrary-scale rendering and any-ratio progressive pruning within a single model, enabling continuous rate--quality control beyond discrete layered bitstreams.

\section{Preliminaries}
\label{sec:method}
\textbf{2D Gaussian Splatting (2DGS):} 3DGS employs anisotropic 3D Gaussian primitives to synthesize novel views from sparse input images, capturing continuous variations in 3D space. When the target is 2D video frame generation, however, the representation is overly expressive, since the depth ordering and view-dependent effects are no longer needed. Motivated by this observation, we adopt the 2D Gaussian formulation~\cite{zhang2024gaussianimage} for efficient rasterization of 2D frames. 
With this formulation, each primitive is parameterized by its 2D position $\boldsymbol{\mu}\in\mathbb{R}^2$, 2D covariance matrix $\Sigma\in\mathbb{R}^{2\times 2}$, and color coefficient vector $\boldsymbol{c}\in\mathbb{R}^3$. The 2DGS implementation ~\cite{zhang2024gaussianimage} omits $\alpha$-blending and opacity, as the depth ordering is unnecessary for frame regression, allowing $\boldsymbol{c}$ to implicitly absorb the opacity effects. 

To ensure numerical stability while retaining an anisotropic representation, $\Sigma$ is parameterized via using the rotation-scaling decomposition:
\begin{equation}
\Sigma = R(\theta)\,\mathrm{diag}(s_x^2, s_y^2)\,R(\theta)^\top,\;
R(\theta)=
\begin{bmatrix}
\cos\theta & -\sin\theta\\
\sin\theta & \cos\theta
\end{bmatrix}.
\end{equation}
where $s_x,s_y>0$ are the principal-axis standard deviations and $\theta\in\mathbb{R}$ is the in-plane rotation. During rasterization, the color vector at pixel $i$ is obtained by summing contributions from all primitives:
\begin{equation}
\boldsymbol{c}_i=\sum_{n\in\mathcal{N}} \boldsymbol{c}_n\,
\exp\!\left(-\tfrac12\,\boldsymbol{d}_n^\top \Sigma_n^{-1}\boldsymbol{d}_n\right),
\end{equation}
where $\boldsymbol{d}_n\in\mathbb{R}^2$ denotes the displacement between the pixel center and the Gaussian center.

\textbf{Nyquist Sampling Criterion:}
When rendering on a discrete pixel grid, content with frequencies exceeding half of the grid's inherent Nyquist frequency results in aliasing artifacts. For a target resolution downsampled by the factor $s$, the effective pixel spacing increases by $s$, and the corresponding half Nyquist angular frequency becomes $\omega_N=\pi/s$. A common approach for reducing aliasing is to ensure sufficient attenuation of signal energy near or above $\omega_N$ by applying low-pass filtering before downsampling. This will be used in Section~\ref{scale-aware} to gate Gaussian primitives across scales based on their pixel-domain footprints, so that coarse renderings preferentially retain smoother Gaussians.

\textbf{D-optimality:}
D-optimality is a classical experimental-design criterion that selects a subset of measurements to maximize the determinant of the resulting information (or Gram) matrix, equivalently maximizing the log-determinant $\log\det(\cdot)$. Intuitively, this favors subsets whose induced features are both strong and diverse, since $\det(\cdot)$ grows when selected elements have large “energy” but remain weakly correlated. In our setting, choosing a $K$-subset of Gaussians can be viewed as selecting $K$ induced Gaussian kernels, so D-optimal selection naturally captures the goal of keeping influential primitives while avoiding redundant overlap.



\begin{figure*}
    \centering
    \includegraphics[width=\linewidth]{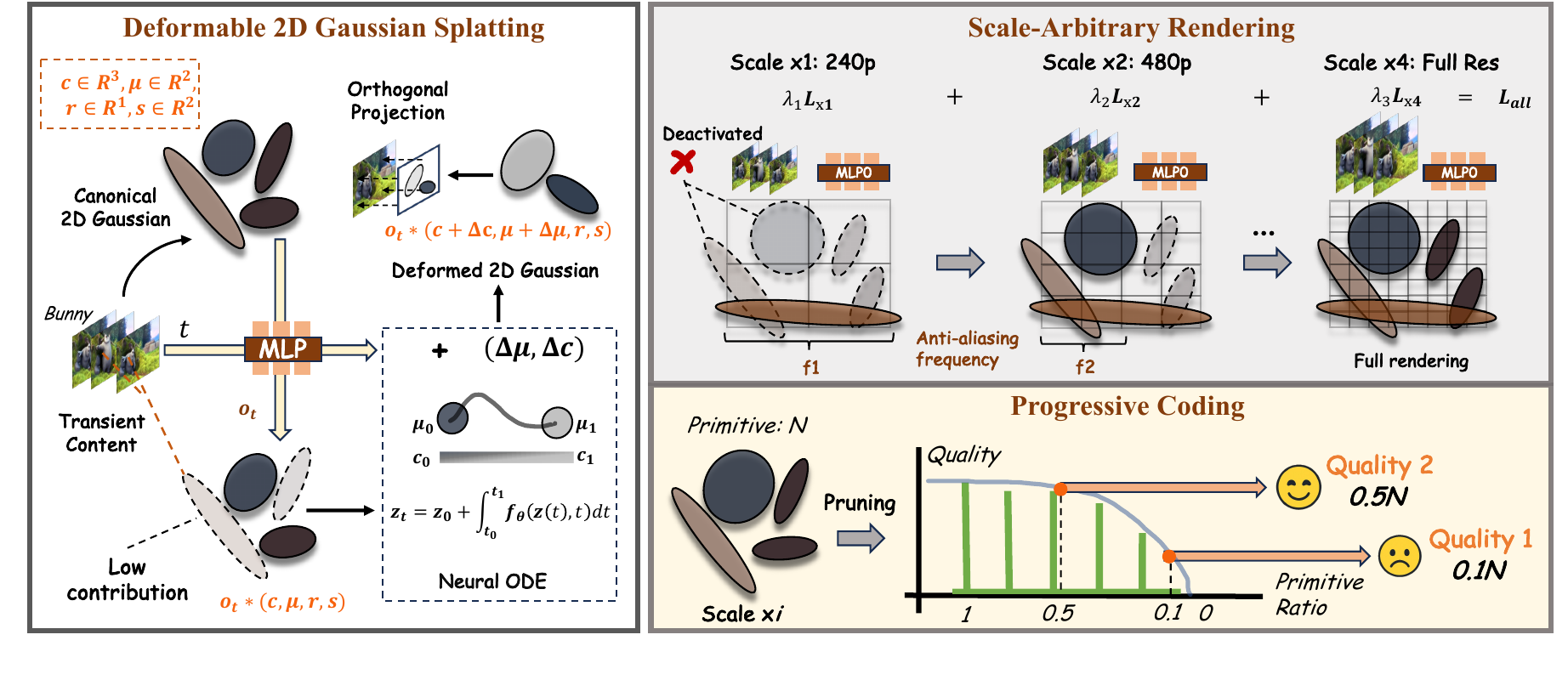}
    \vspace{-0.8cm}
    \caption{Overview of D2GV-AR. Left: deformable 2D Gaussian splatting framework. Right: top, a toy example of scale-aware grouping for rendering at three resolutions; bottom, progressive pruning for continuous rate–quality control.}
    \label{overview}
    \vspace{-0.5cm}
\end{figure*}

\section{Deformable 2DGS}

\subsection{Deformation Field}
We model a video as a \emph{time-varying 2D Gaussian field} subject to continuous deformation. Instead of learning an independent set of Gaussians for every frame, we maintain a canonical Gaussian set $\mathcal{G}_0$, and describe the entire sequence through a mapping that transports $\mathcal{G}_0$ to the frame at timestamp $t$. Mathematically, let $\mathcal{V}=\{V_1,\dots,V_T\}$ be the video frame set and $\mathcal{G}_t$ denote the 2D Gaussian parameters used to synthesize frame $V_t$. We define a deformation field $f_\theta(\cdot,t)$ and generate the time-dependent Gaussian set using
\begin{equation}
\mathcal{G}_t = \mathcal{G}_0 \oplus f_\theta(\mathcal{G}_0,t),    
\end{equation}
where $\oplus$ denotes applying the produced offsets $f_\theta(\mathcal{G}_0,t)$ to the canonical Gaussian parameters. The rendered frame can then be obtained through differentiable rasterization
\begin{equation}
\hat V_t = \mathcal{R}(\mathcal{G}_t) = \mathcal{R}\!\left(\mathcal{G}_0 \oplus f_\theta(\mathcal{G}_0,t)\right),
\end{equation}
with $\mathcal{R}(\cdot)$ being the rasterization operator. Since $\mathcal{R}(\cdot)$ is differentiable, gradients from an image-space reconstruction loss on $\hat V_t$ can propagate through the deformation field back to both the canonical representation $\mathcal{G}_0$ and deformation neural network parameters $\theta$, enabling end-to-end training.

\subsection{Implementations}
We initialize the canonical representation $\mathcal{G}_0$ by optimizing a set of 2D Gaussians over all frames in a sequence, yielding a prior that captures shared content. As the deformation is described using the neural ODE, modeling continuous-time motion without any explicit grid is possible (see Fig. \ref{overview}).

Given time $t$ and a Gaussian center $\boldsymbol{\mu}$, we maintain a latent deformation state $\mathbf{s}(t)$ whose dynamics are predicted by a lightweight MLP
\begin{equation}
\frac{d\mathbf{s}(t)}{dt}=\text{MLP}\big(\gamma(\boldsymbol{\mu}),\,\gamma(t)\big),
\end{equation}
where $\gamma(\cdot)$ denotes a standard sinusoidal positional encoding (applied to both $\boldsymbol{\mu}$ and $t$). Starting from $\mathbf{s}(t_0)$, we obtain $\mathbf{s}(t)$ by integrating over $[t_0,t]$. We then decode deformation offsets with a fully-connected (FC) layer head
\begin{equation}
(d\boldsymbol{\mu},\,d\boldsymbol{c})=\mathrm{FC}\big(\mathbf{s}(t)\big).
\end{equation}
To account for non-smooth appearance changes, we predict an extra per-frame opacity gate with a separate head
\begin{equation}
o_t=\text{sigmoid}\Big(\mathrm{FC}\big(\gamma(\boldsymbol{\mu}),\,\gamma(t)\big)\Big),\qquad o_t\in(0,1).
\end{equation}
This gate is decoupled from the deformation $s(t)$, allowing it to be used to suppress primitives that are irrelevant at specific timestamps. The 2DGS deformation is thus
\begin{equation}
\boldsymbol{\mu}'=\boldsymbol{\mu}+d\boldsymbol{\mu},\qquad
\boldsymbol{c}'=o_t\big(\boldsymbol{c}+d\boldsymbol{c}\big).
\end{equation}
Covariances of the Gaussians are kept intact, since empirical studies showed that deforming them can make the canonical grouping in Section 5.1 required for arbitrary-scale rendering unstable.

\section{Progressive Arbitrary-scale Representation}
The explicit Gaussian representation enables progressive multi-resolution rendering based on a single set of primitives. We explore this together with (i) \emph{scale-aware Gaussian grouping} to support \emph{arbitrary-scale} rendering from one representation, and (ii) \emph{progressive} coding to enable \emph{continuous} rate-quality control. Notably, multiscale training also improves the quality of progressive decoding.

\subsection{Scale-aware Gaussian Grouping}
\label{scale-aware}
We aim at rendering at an arbitrary target resolution by selecting a scale-dependent subset of canonical Gaussians, following a criterion similar to Nyquist sampling theory. For a downsampling ratio $r$, we retain the group
\begin{equation}
\mathcal{G}^{(r)}_0 \triangleq \{\, n \mid \sigma_{\max,n} \ge \beta\, r \,\},
\end{equation}
where $\sigma_{\max,n}$ is the pixel-domain footprint of Gaussian $n$ and $\beta$ is a user-defined threshold. With a single $\beta$, this grouping rule results in a nested hierarchy, i.e., 
\begin{equation}
\mathcal{G}_0^{(r_2)} \subseteq \mathcal{G}_0^{(r_1)} \quad \text{for } r_2 > r_1.
\end{equation}
In other words, lower-resolution decoding always uses fewer primitives. The footprint $\sigma_{\max,n}$ is estimated from $\mathcal{R}(\Sigma)$, the rasterized version of the covariance of Gaussian $n$, using
\begin{equation}
\sigma_{\max} \triangleq \sqrt{\lambda_{\max}\!\big(\mathcal{R}(\Sigma)\big)},
\end{equation}
with $\lambda_{\max}(\cdot)$ denoting the biggest eigenvalue. Note again that the covariances of the Gaussians $\Sigma$ are fixed so that the footprint of each Gaussian in the prior $\mathcal{G}_0$ is invariant, which stabilizes the grouping operation across frames.

\begin{theorem}[Nyquist sampling-like criterion]\label{thm:nyquist_gaussian}
Let $g(x)=\exp\!\big(-x^2/(2\sigma^2)\big)$ with frequency response $|\hat g(\omega)|\propto \exp\!\big(-\tfrac{1}{2}\sigma^2\omega^2\big)$.
On a grid with spacing $\Delta$ (half of the Nyquist frequency $\omega_N=\pi/\Delta$), a sufficient anti-aliasing condition
$|\hat g(\omega_N)|\le \varepsilon$ holds if
\[
\sigma \ge \beta(\varepsilon)\,\Delta,
\qquad
\beta(\varepsilon)\triangleq \frac{\sqrt{2\ln(1/\varepsilon)}}{\pi}.
\]
\end{theorem}
See Appendix \ref{Nyquist} for a proof.

\textbf{Remark}: Downsampling by a ratio of $r$ means sampling on a rougher pixel grid with spacing $\Delta \propto r$.
Theorem~\ref{thm:nyquist_gaussian} therefore implies a footprint threshold $\sigma_{\max}\propto r$,
supporting our grouping criterion design with a single factor $\beta$.

\subsection{Progressive Coding}
At scale $r$, we decode a $K$-subset $S\subseteq \mathcal{G}_0^{(r)}$ for continuous rate--quality control. We prioritize primitives with larger screen-space coverage, while penalizing redundancy when multiple Gaussians overlap on the same pixels. This criterion admits a principled formulation as a D-optimal subset selection problem over the induced Gaussian kernels. We formalize this below.
\begin{definition}[D-optimal pruning]\label{def:dopt_prune}
{\small\itshape
Let $N_r\triangleq|\mathcal{G}_0^{(r)}|$ and index $\mathcal{G}_0^{(r)}$ by $\{1,\ldots,N_r\}$.
For primitive $n$, define the induced 2D Gaussian kernel
$\phi_n(\boldsymbol{x})=\exp\!\big(-\tfrac12(\boldsymbol{x}-\boldsymbol{\mu}_n)^\top\Sigma_n^{-1}(\boldsymbol{x}-\boldsymbol{\mu}_n)\big)$.
Let $G\in\mathbb{R}^{N_r\times N_r}$ be the Gram matrix with
$G_{ij}\triangleq\langle \phi_i,\phi_j\rangle=\int_{\mathbb{R}^2}\phi_i(\boldsymbol{x})\phi_j(\boldsymbol{x})\,d\boldsymbol{x}$.
Given a budget $K$, we select
\[
S^\star \in \arg\max_{S\subseteq\mathcal{G}_0^{(r)},\,|S|=K}\ \log\det\!\big(G[S]\big),
\]
where $G[S]$ is the principal submatrix indexed by $S$.
}
\end{definition}
We factorize $G$ as $G=DRD$ with
$D=\mathrm{diag}(\sqrt{G_{11}},\ldots,\sqrt{G_{N_rN_r}})$ and $R=D^{-1}GD^{-1}$.
Then
\begin{equation}
\label{decompose}
\log\det\!\big(G[S]\big)=\sum_{n\in S}\log G_{nn}+\log\det\!\big(R[S]\big),
\end{equation}
where the first term captures footprint coverage, while $\log\det(R[S])$ is a coupled term that measures overall overlap among selected primitives. Since it cannot be decomposed into per-primitive contributions, we approximate it with an additive quadratic surrogate.
\begin{lemma}[Additive quadratic surrogate]\label{lem:overlap_surrogate}
For any $S$ with $|S|=K$, define $A_S \triangleq R[S]-I_K$ and assume $\|A_S\|_2\le \alpha<1$.
Then there exists a remainder term $\Delta(S)$ such that
{\small
\begin{align*}
\log\det\!\big(R[S]\big)
&= -\frac12\!\sum_{\substack{i,j\in S\\ i\neq j}} R_{ij}^2 \;+\; \Delta(S),\\
|\Delta(S)|
&\le \frac{\alpha}{3(1-\alpha)}
\sum_{\substack{i,j\in S\\ i\neq j}} R_{ij}^2 .
\end{align*}
}
\end{lemma}
See Appendix~\ref{proof53} for a proof.\\
Since $G_{nn}=\|\phi_n\|_2^2=\pi s_{x,n}s_{y,n}$ (Appendix~\ref{area}), replacing the second term in Eq. (\ref{decompose}) with the additive surrogate in Lemma~\ref{lem:overlap_surrogate} reduces the problem to selecting a subset $S$ of $K$ primitives by
\begin{equation}
S^\star \in \arg\max_{\substack{S\subseteq \mathcal{N}\\ |S|=K}}
\left\{\sum_{n\in S}\log(\pi s_{x,n}s_{y,n})
-\frac{1}{2}\!\sum_{\substack{i,j\in S\\ i\neq j}}R_{ij}^2\right\}.
\end{equation}
For fast one-shot scoring, we approximate the redundancy sum with local neighborhoods:
let $\mathcal{K}_n$ be the $k$ nearest neighbors of primitive $n$ and set
$\rho_n\triangleq\sum_{j\in\mathcal{K}_n}R_{nj}^2$, yielding the ranking score
$\mathrm{score}_n=(s_{x,n}s_{y,n})\exp(-\lambda\rho_n)$ followed by top-$K$ selection, where $\lambda_s\ge 0$ controls the coherence penalty.

\subsection{Training Objective}
Following the specifications of the conventional codecs, we adopt GoP-level training with a fixed GoP size, since long-range deformations across scene changes are difficult to model reliably. For each GoP, we minimize the distortion between the original and reconstructed frames over the entire segment. During training, we supervise the reconstruction over a set of resolutions $\mathcal{R}_{\text{train}}$ as the multi-task form. The objective is
\begin{equation}\label{eq:opt}
\mathcal{L} = \sum_{r\in\mathcal{R}_{\text{train}}} \lambda_r \Big(\mathcal{L}_{\text{PSNR}}^{(r)} + \lambda_s \mathcal{L}_{\text{SSIM}}^{(r)}\Big),
\end{equation}
where $\lambda_r$ weights each scale, and $\lambda_s$ weights the SSIM term. $\mathcal{L}_{\text{PSNR}}^{(r)}$ and $\mathcal{L}_{\text{SSIM}}^{(r)}$ are computed between the reconstructed version and ground truth at scale $r$. After training, the model can render at arbitrary resolutions with any-ratio pruning.

\begin{figure}[t]
    \centering
    \includegraphics[width=\linewidth]{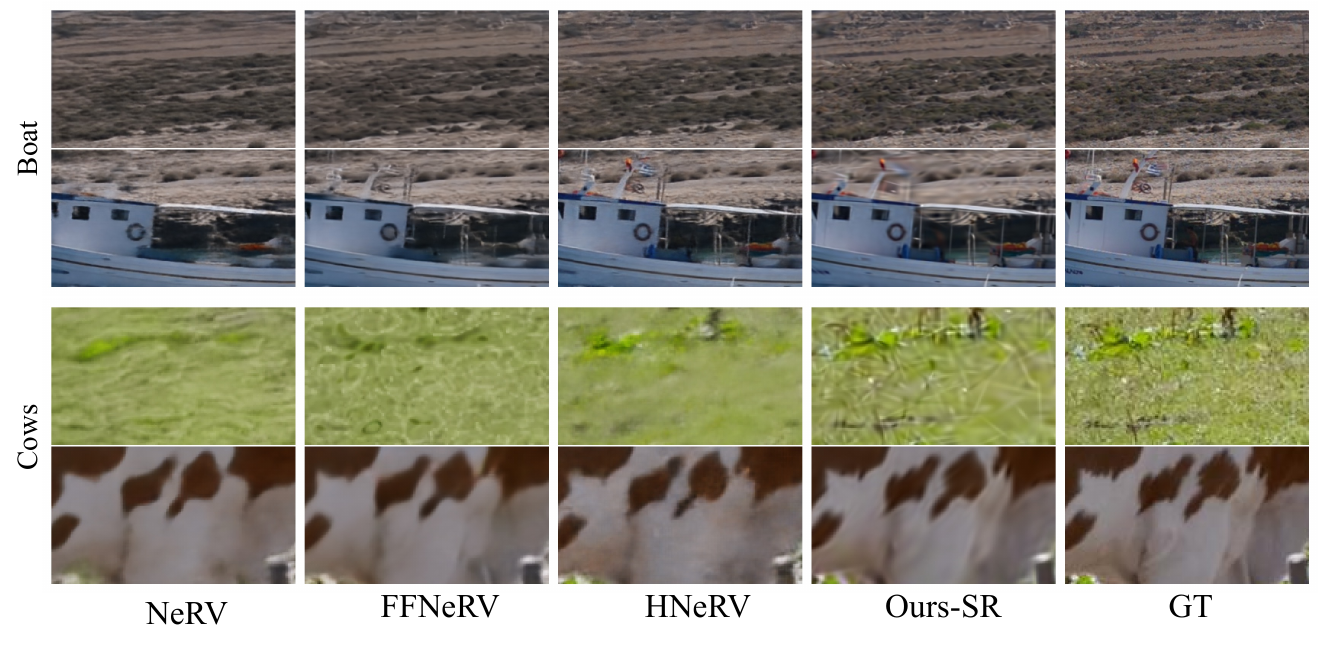}
    \caption{Single-resolution visual comparions. Zoom in to see details.}
    \vspace{-0.4cm}
    \label{singlescalevis}
\end{figure}

\section{Experiment}
\textbf{Dataset. }We evaluate our approach on three datasets: Big Buck Bunny, UVG \cite{mercat2020uvg}, and DAVIS \cite{perazzi2016benchmark}. Big Buck Bunny is obtained from scikit-video and contains 132 frames at 720p resolution. UVG comprises seven sequences{\tiny\footnote{Bosphorus, Beauty, SetGo, Bee, Yacht, Jockey, Shake.}}, and we choose six sequences from DAVIS{\tiny\footnote{Train, Camel, Bmx-tree, Blackswan, Cow, Boat.}}. For fair comparison with prior work, we follow the evaluation protocol of E-NeRV \cite{li2022nerv}: we resize UVG and DAVIS to 720p, and apply $4\times$ temporal downsampling to UVG.\\
\textbf{Evaluation. }We evaluate reconstruction quality using peak signal-to-noise ratio (PSNR) \cite{PSNR} and multi-scale structural similarity (MS-SSIM) \cite{wang2003multiscale}. We report rendering speed in frames per second (FPS) and compression performance in bits per pixel (BPP). Unless otherwise specified, all model parameters are stored in FP32. We compare our method against INR- and GS-based approaches under both single-scale and multi-scale settings. For single-scale evaluation, We benchmark our method against NeRV \cite{chen2021nerv}, E-NeRV \cite{li2022nerv}, FFNeRV \cite{lee2023ffnerv}, and HNeRV \cite{chen2023hnerv}, and include a deformable 3DGS, D-3DGS \cite{yang2024deformable}. These methods are designed to operate at a single scale and must be retrained for other scales. For multi-resolution evaluation, we compare with multi-resolution NeRV \cite{NeRV_GitHub_2021} and SNeRV \cite{wei2024snerv}. Baseline details are provided in Appendix \ref{baseline}.\\
\textbf{Implementation details. }
We set the GoP size to 10.
The Neural ODE is parameterized by a one-layer MLP with hidden width 156, using positional/temporal encodings with 10/6 frequency bands for 2D coordinates and timestamps.
We integrate the ODE with a fixed-step fourth-order Runge--Kutta solver (\texttt{torchdiffeq}).
We set the SSIM loss weight $\lambda_s{=}0.3$ and the coherence penalty weight $\lambda_r{=}3$.
For multi-scale training, the original scale is weighted by 8 and all other scales by 1, as lowering the original-scale weight degrades overall BD-rate.
All experiments run on a single NVIDIA RTX 3090 GPU.
Additional details about hyperparameter setting are provided in Appendix~\ref{experimental}.

\begin{table}[t]
\centering
\caption{Average video regression results (PSNR/MS-SSIM) on two model sizes.}
\label{tab:avg_regression}
\setlength{\tabcolsep}{6pt}
\renewcommand{\arraystretch}{1.12}

\resizebox{\linewidth}{!}{%
\small
\begin{tabular}{@{} l l c c c c @{}}
\toprule
\textsc{Dataset} & \textsc{Method} &
\multicolumn{2}{c}{\textsc{Small}} &
\multicolumn{2}{c}{\textsc{Large}} \\
\cmidrule(lr){3-4}\cmidrule(lr){5-6}
& & \textsc{PSNR}$\uparrow$ & \textsc{MS-SSIM}$\uparrow$ & \textsc{PSNR}$\uparrow$ & \textsc{MS-SSIM}$\uparrow$ \\
\midrule

\multirow{6}{*}{\textsc{Davis}}
& NeRV     & 26.08 & 0.8699 & 26.82 & 0.8902 \\
& E-NeRV   & 27.12 & 0.9044 & 28.55 & 0.9232 \\
& FFNeRV   & 28.20 & \underline{0.9242} & 29.73 & \underline{0.9484} \\
& HNeRV    & \underline{29.31} & 0.9115 & \underline{31.53} & 0.9459 \\
& D-3DGS   & 24.54 & 0.8385 & 26.48 & 0.8808 \\
& Ours-SR  & \textbf{29.82} & \textbf{0.9457} & \textbf{31.87} & \textbf{0.9668} \\
\addlinespace[2pt]
\midrule

\multirow{6}{*}{\textsc{UVG}}
& NeRV     & 33.04 & 0.9486 & 33.40 & 0.9521 \\
& E-NeRV   & 33.48 & \underline{0.9675} & 35.02 & 0.9712 \\
& FFNeRV   & \underline{35.17} & \textbf{0.9701} & 35.73 & 0.9720 \\
& HNeRV    & \textbf{35.59} & 0.9607 & \textbf{37.18} & \textbf{0.9765} \\
& D-3DGS   & 26.27 & 0.8868 & 28.54 & 0.9124 \\
& Ours-SR  & 35.03 & 0.9663 & \underline{35.83} & \underline{0.9728} \\
\bottomrule
\end{tabular}%
}
\end{table}

\begin{figure}[t]
    \centering
    \includegraphics[width=0.95\linewidth]{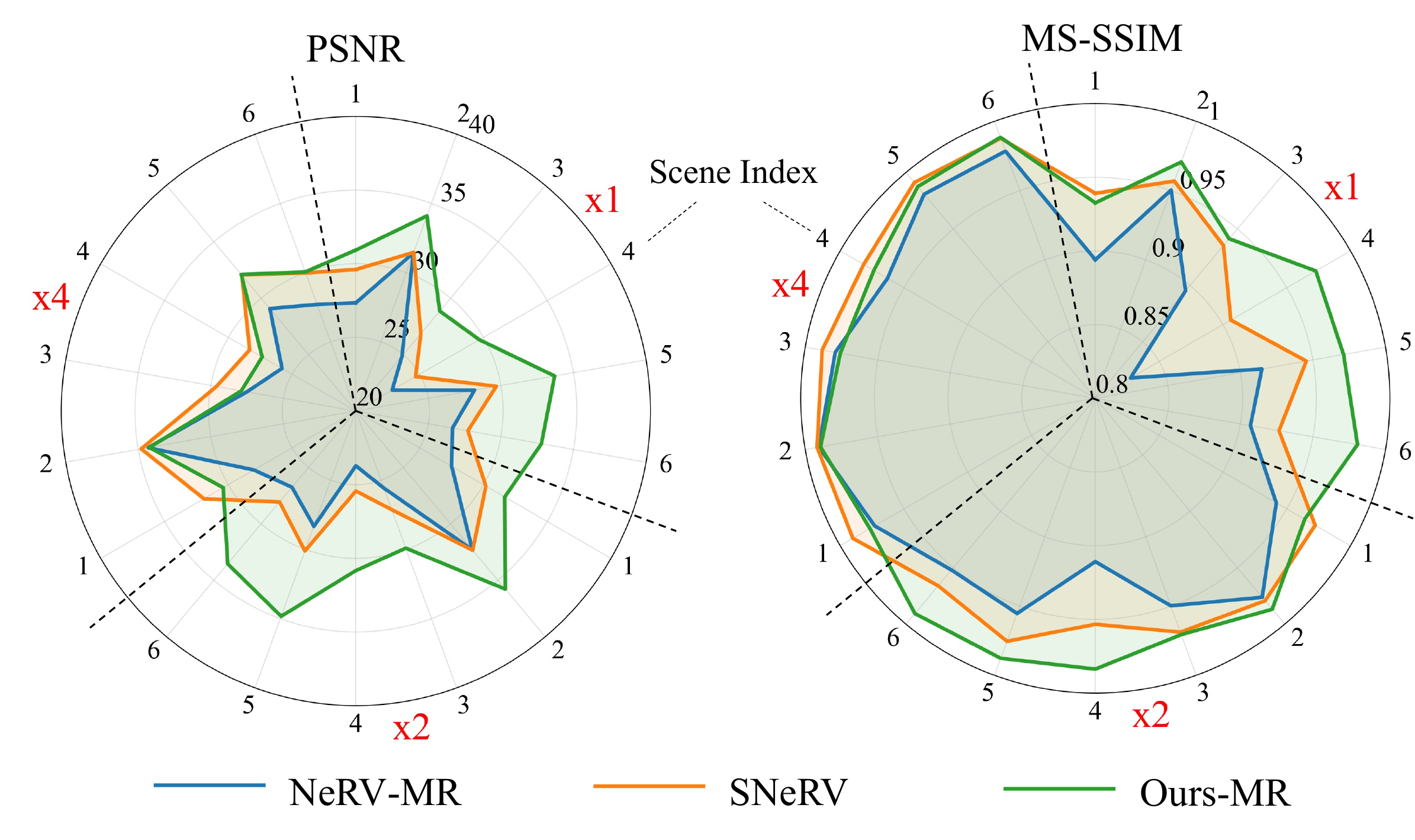}
    \caption{Qualitative multi-resolution comparisons on six DAVIS scenes (model in large size setting).}
    \label{radar}
\end{figure}

\begin{figure*}
    \centering
    \includegraphics[width=0.95\linewidth]{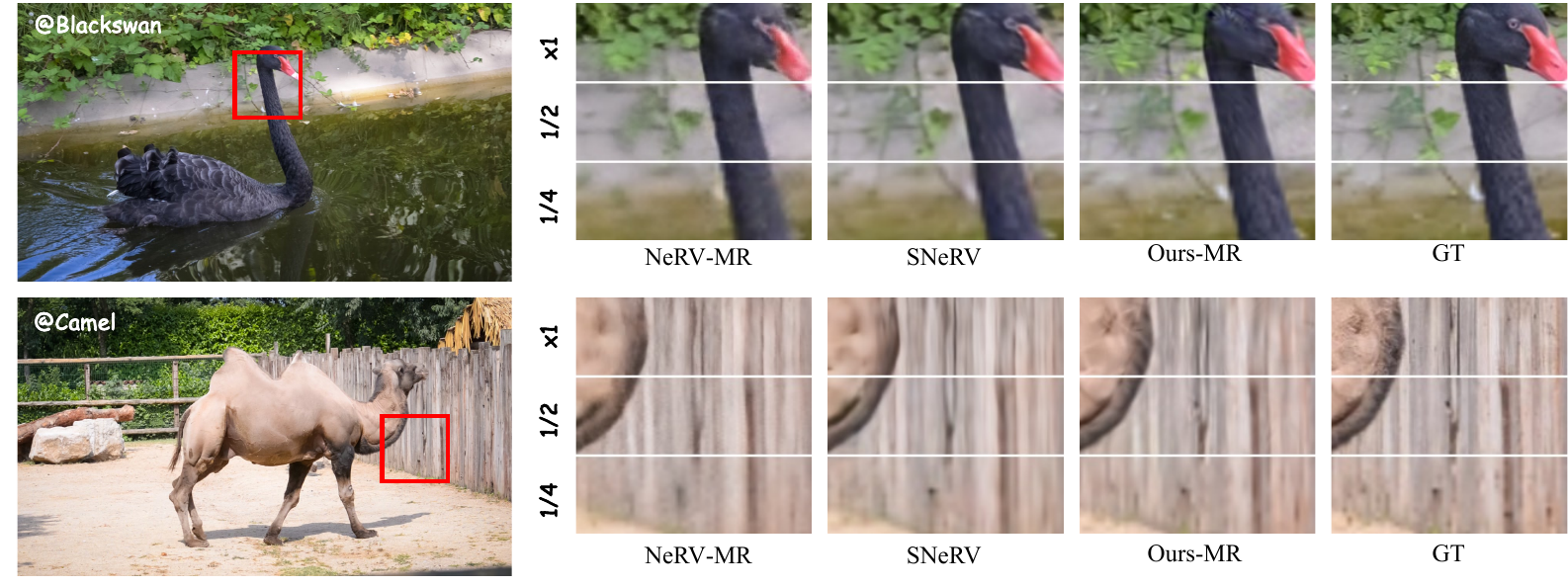}
    \caption{Multi-resolution visual comparisons. The original scale x1, downscale x2 and x4 are displayed.}
    \vspace{-0.6cm}
    \label{multiscalevis}
\end{figure*}

\begin{table}[t]
\centering
\caption{PSNR and primitive ratio on Bunny (30k primitives) vs. scales. Training scales are labeled in black.}
\label{tab:scale_psnr_maskratio_one}
\setlength{\tabcolsep}{4pt}
\renewcommand{\arraystretch}{1.08}
\resizebox{0.8\linewidth}{!}{%
\small
\begin{tabular}{@{}lccccc@{}}
\toprule
\textsc{Scale} &
\textbf{\textsc{Full}} & $\times$1.5 & \textbf{$\times$2} & $\times$2.5 & $\times$3 \\
\midrule
\textsc{PSNR (dB)} & 37.39 & 38.75 & 38.31 & 37.14 & 36.12 \\
\textsc{Primitive Ratio (\%)} & 100.00 & 95.00 & 86.68 & 81.97 & 72.29 \\
\midrule
\textsc{Scale} &
$\times$3.5 & \textbf{$\times$4} & $\times$4.5 & $\times$5 & \textbf{$\times$8} \\
\midrule
\textsc{PSNR (dB)} & 35.35 & 36.62 & 33.52 & 34.70 & 32.61 \\
\textsc{Primitive Ratio (\%)} & 67.63 & 58.64 & 54.33 & 47.06 & 18.21 \\
\bottomrule
\end{tabular}%
}
\end{table}

\subsection{Main Results}
\begin{figure}[t]
    \centering
    \includegraphics[width=0.99\linewidth]{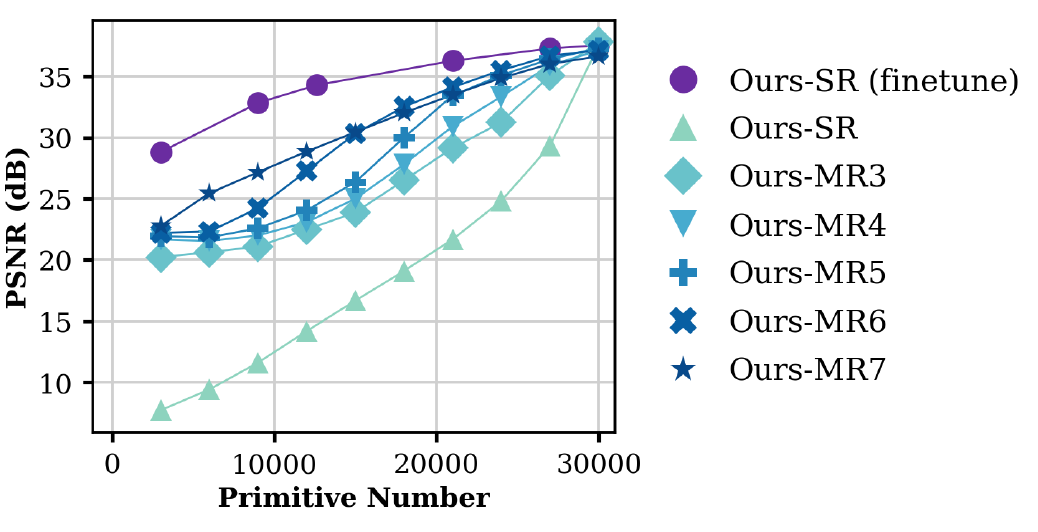}
    \caption{Progressive coding performance vs. training scales on Bunny dataset.}
    \label{progressive}
\end{figure}

\begin{figure}[t]
    \centering
    \includegraphics[width=0.9\linewidth]{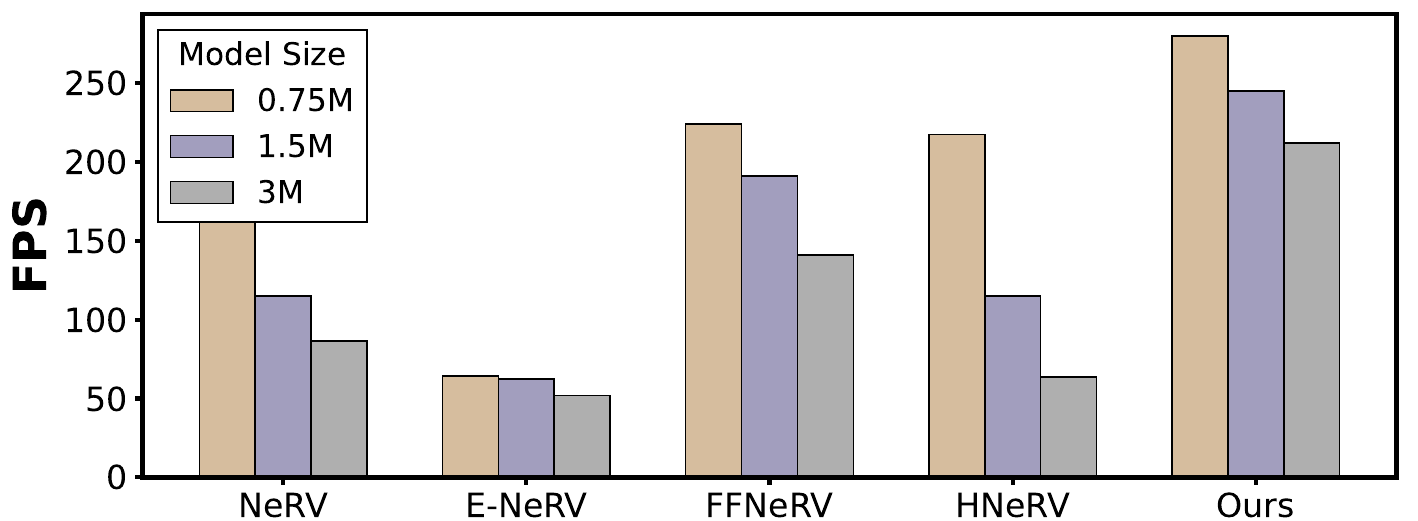}
    \caption{Comparisons on rendering speed, evaluated on the Bunny dataset.}
    \label{fps}
    \vspace{-0.5cm}
\end{figure}

We evaluate our method under both single- and multi-resolution settings, covering regression quality metrics, progressive coding, and efficiency. Additional results are in Appendix~\ref{moreresults}.\\
\textbf{Single-resolution. }
Table~\ref{tab:avg_regression} reports PSNR and MS-SSIM under two model-size budgets (in 26k and 46k primitives) on UVG and DAVIS. Our method performs best on DAVIS and remains competitive on UVG. On DAVIS, under the small budget, our model improves the average PSNR by $+1.7\%$ over HNeRV and the average MS-SSIM by $+2.3\%$ over FFNeRV; similar gains are observed under the large budget. D-3DGS exhibits a clear gap under the same budget, largely due to the higher storage overhead of 3D Gaussian parameterization, which reduces the effective capacity for 2D reconstruction. Qualitative results in Fig.~\ref{singlescalevis} further confirm that our method preserves finer structures and recovers sharper details. \\
\textbf{Multi-resolution. }
In the multi-resolution setting, all methods are trained at four scales ($\times1$, $\times2$, $\times4$, $\times8$) with multi-task supervision for a fair comparison. We report results at $\times1$, $\times2$, and $\times4$ in Fig.~\ref{radar}. Our model forms the outer envelope on most scene--scale axes for both PSNR and MS-SSIM, indicating more consistent multi-scale quality than NeRV-MR and SNeRV, with only a few exceptions at $\times4$. The few $\times4$ exceptions mainly stem from our resolution re-weighting, where the original scale is upweighted to optimize overall BD-rate and the $\times4$ branch can be slightly under-trained. Beyond the trained branches, our representation also supports unseen-scale rendering; Table~\ref{tab:scale_psnr_maskratio_one} shows an example on \textit{Bunny} dataset.\\
\textbf{Progressive Coding. }
To compare pruning criteria, we build an opacity-based baseline by learning per-primitive opacity for ranking (opacity-based). Table~\ref{tab:bd_opacity_anchor} shows that area-based ranking yields a better rate--quality trade-off than radii, suggesting footprint is a more reliable utility proxy. Adding our coherence penalty further improves BD-rate/BD-PSNR, indicating that effective pruning should explicitly penalize overlap rather than only favor large splats. We further study how multi-scale training affects pruning behavior in Fig.~\ref{progressive}. Post-hoc finetuning after pruning (finetune) provides a quality upper bound. Training with an increasing set of resolutions consistently narrows the gap to this bound, indicating that multi-scale supervision improves pruning robustness in practice. More results are provided in Appendix \ref{pcmore}.\\
\textbf{Efficiency. }Our method achieves significantly high rendering FPS in the single-resolution Bunny setting (Fig.~\ref{fps}). Notably, our CUDA-based rasterization runs at over $250$ FPS, which is about $1.2\times$ faster than the strongest INR baseline FFNeRV and over $4\times$ faster than E-NeRV under the same budget.


\begin{table}[t]
\caption{BD-rate / BD-PSNR vs. opacity-based pruning.}
\label{tab:bd_opacity_anchor}
\centering
\small
\resizebox{0.9\linewidth}{!}{
\begin{tabular}{lcc}
\toprule
\textsc{Method (vs. Opacity-based)} & \textsc{BD-rate} $\downarrow$ & \textsc{BD-PSNR} $\uparrow$ \\
\midrule
\textsc{Radii} & $-9.18\%$  & $+1.56$ \\
\textsc{Area} & $-24.33\%$ & $+4.25$ \\
\textsc{Area+Coherence penalty} & $\textbf{-25.21\%}$ & $\textbf{+4.52}$ \\
\bottomrule
\end{tabular}}
\vspace{-0.5cm}
\end{table}

\begin{table*}[t]
\centering
\caption{Video interpolation results with train/test split. PSNR (dB) is reported as Train/Interp frames.}
\label{interpolation}
\setlength{\tabcolsep}{5pt}
\renewcommand{\arraystretch}{1.10}

\resizebox{0.75\textwidth}{!}{%
\small
\begin{tabular}{@{} l c c c c c c c @{}}
\toprule
\textsc{Method} & \textsc{Boat} & \textsc{Camel} & \textsc{Blackswan} & \textsc{Train} & \textsc{Cows} & \textsc{Bmx-tree} & \textsc{Average} \\
\midrule
\textsc{FFNeRV}
& 27.21/16.27 & 21.79/13.89 & 23.84/14.50 & 22.20/15.82 & 20.64/16.59 & 24.14/15.11 & 23.30/15.36 \\
\textsc{Ours w/o NeuralODE}
& 26.59/26.43 & \textbf{23.09}/19.65 & 25.79/20.05 & 23.11/\textbf{20.87} & 19.49/19.31 & \textbf{26.90}/16.88 & 24.16/20.53 \\
\textsc{Ours w/ NeuralODE}
& \textbf{30.17}/\textbf{29.28} & 22.66/\textbf{20.71} & \textbf{25.85}/\textbf{20.79} & \textbf{23.59}/20.04 & \textbf{23.12}/\textbf{22.36} & 20.83/\textbf{16.99} & \textbf{24.37}/\textbf{21.69} \\
\bottomrule
\end{tabular}%
}
\end{table*}

\subsection{Downstream Tasks}
\begin{figure}[t]
    \centering
    \begin{minipage}{0.49\linewidth}
        \centering
        \includegraphics[width=\linewidth]{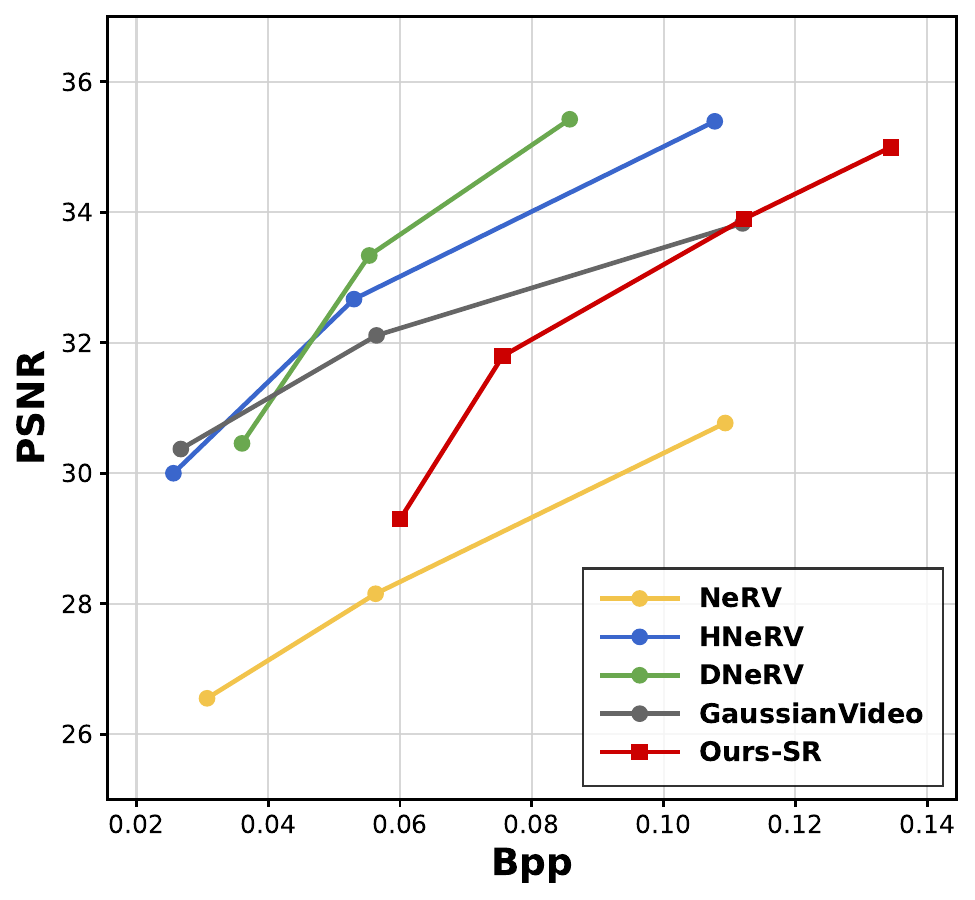}\\
        {\small Left: single-resolution.}
    \end{minipage}\hfill
    \begin{minipage}{0.49\linewidth}
        \centering
        \includegraphics[width=\linewidth]{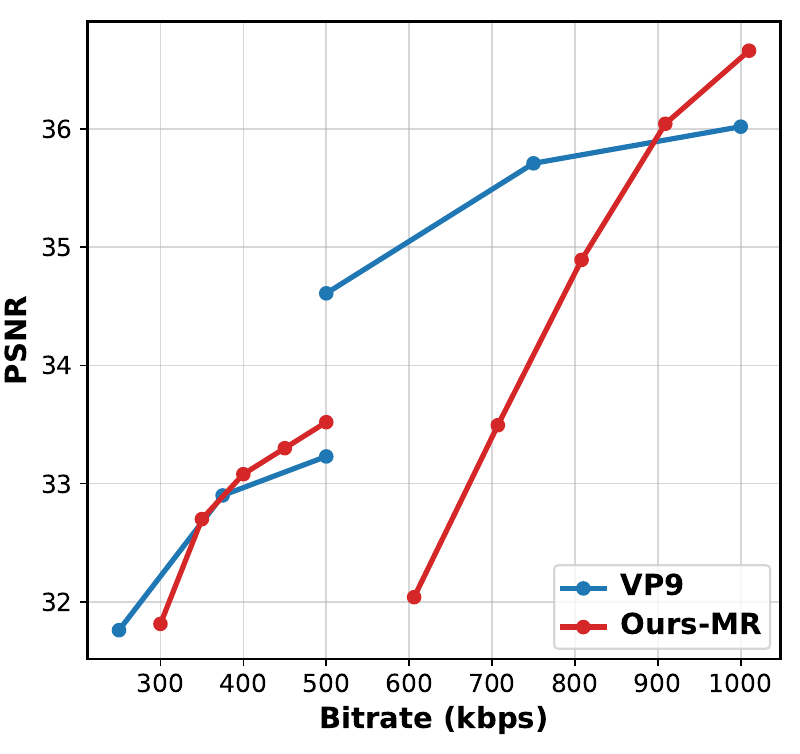}\\
        {\small Right: multi-resolution.}
    \end{minipage}
    \caption{Compression results on Bunny (Results for single-resolution are from \cite{lee2025gaussianvideo}; all reconstructions are upsampled to the original resolution before evaluation).}
    \label{fig:compress}
\end{figure}

\textbf{Video Interpolation.}
In the frame interpolation experiments, we train 0.75M-parameter variants of our method (with and without NeuralODE) and compare against the flow-guided INR baseline FFNeRV. Following the standard split, we fit on odd frames and evaluate interpolation on even frames. According to Table \ref{interpolation}, our approach improves interpolation quality by about $+5\!\sim\!6$~dB over FFNeRV and substantially narrows the train–test gap, indicating better generalization. The results show that neural ODE plays a key role by imposing a continuous deformation constraint.\\
\textbf{Video Compression.}
For video compression comparison, we evaluate both single-resolution and multi-resolution settings. In the multi-resolution setting, lower-resolution reconstructions are upscaled to the original resolution for PSNR computation. We adopt the compression pipeline of \cite{zhang2024gaussianimage} to encode canonical Gaussians for single-resolution coding, and apply quantization-aware fine-tuning only in the multi-resolution setting to enable continuous pruning. Results are summarized in Fig.~\ref{fig:compress}. We compare multiscale rate–distortion performance against the SVC codec VP9 \cite{vp9bitstream2016} under 500 kbps and 1000 kbps bitrate budgets. Our method is comparable to INR baselines but still behind VP9, likely because VP9 relies on finite scalability layers ($\leq$ 5), whereas ours enables continuous quality control via fine-grained pruning.\\

\begin{figure}[t]
    \centering
    \includegraphics[width=\linewidth]{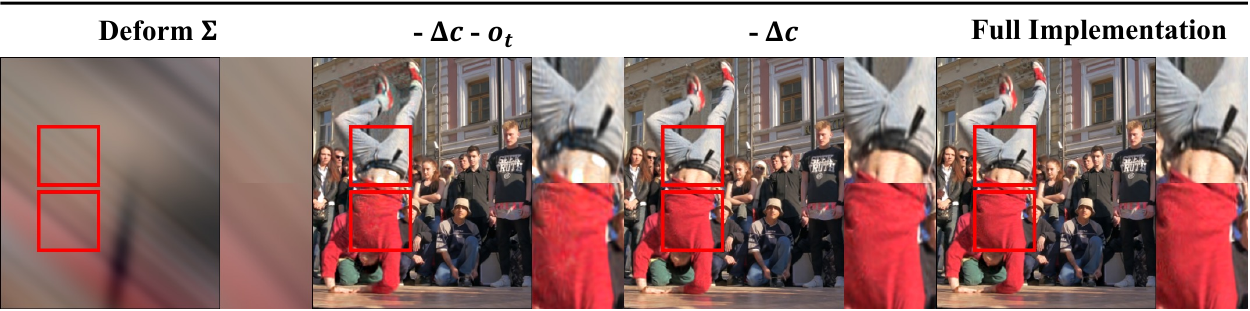}
    \caption{Ablation on deformation components.}
    \label{ablationvis}
\end{figure}

\begin{table}[t]
\centering
\caption{Ablation results reported as percentage differences (\%) w.r.t.\ Ours (w/ 2D Gaussian). Higher is better.}
\label{tab:ablation_percent}
\setlength{\tabcolsep}{5pt}
\renewcommand{\arraystretch}{1.12}

\resizebox{0.7\linewidth}{!}{%
\small
\begin{tabular}{@{} l c c c @{}}
\toprule
\textsc{Setting} & \textsc{PSNR}$\uparrow$ & \textsc{MS-SSIM}$\uparrow$ & \textsc{FPS}$\uparrow$ \\
\midrule
w/o Coarse Training & -4.58  & -0.69  & -14.67 \\
w/ L2               & -9.68  & -2.17  & -21.28 \\
w/ SSIM             & -7.88  & -1.80  & -18.59 \\
w/ L2+SSIM          & +0.00  & +0.00  & +0.00 \\
\midrule
$-\Delta c - o_t$   & -2.15  & -0.37  & +3.93 \\
$-\Delta c$         & -0.63  & -0.22  & -11.57 \\
\midrule
w/ 3D Gaussian      & -20.34 & -5.55  & -42.98 \\
\textbf{Ours (w/ 2D Gaussian)} & \textbf{+0.00} & \textbf{+0.00} & \textbf{+0.00} \\
\bottomrule
\end{tabular}%
}
\end{table}

\begin{table}[t]
\centering
\caption{Ablation on scale-aware grouping and multiscale training. PSNR is reported in dB.}
\label{tab:ablation_sg_mt_psnr}
\setlength{\tabcolsep}{6pt}
\renewcommand{\arraystretch}{1.12}

\resizebox{0.95\linewidth}{!}{%
\small
\begin{tabular}{@{} c c c c c @{}}
\toprule
\multicolumn{2}{c}{\textsc{Ablation Items}} & \multicolumn{3}{c}{\textsc{PSNR}$\uparrow$ (dB)} \\
\cmidrule(lr){1-2}\cmidrule(lr){3-5}
\textsc{Scale-aware Grouping} & \textsc{Multiscale Training} & \textsc{x1} & \textsc{x2} & \textsc{x4} \\
\midrule
\xmark & \xmark & 37.64 & 27.59 & 20.37 \\
\cmark & \xmark & 37.64 & 29.72 & 19.32 \\
\cmark & \cmark & 36.43 & 38.21 & 36.21 \\
\bottomrule
\end{tabular}%
}
\end{table}

\subsection{Ablation study}
We ablate training strategy, deformation components, and multi-scale design in this section. More ablations can be found at Appendix \ref{Ablation Study}.\\
\textbf{Training strategys.}
Removing the coarse-to-fine schedule degrades reconstruction quality and slows down decoding, indicating that this schedule is necessary for both training stability and runtime efficiency. Moreover, optimizing a single objective (either $\ell_2$ or SSIM alone) consistently underperforms the combined loss, indicating that accurate reconstruction benefits from jointly enforcing pixel fidelity and structural consistency.\\
\textbf{Deformation components.}
Table~\ref{tab:ablation_percent} isolates the deformation terms. Removing both $\Delta c$ and $o_t$ leads to a clear fidelity drop, while removing $\Delta c$ alone results in a smaller yet consistent degradation. These effects are also visible in Fig.~\ref{ablationvis}. In contrast, enabling covariance deformation can destabilize the learned grouping and even prevent convergence, supporting our choice to keep it fixed for stable Gaussian grouping. These trends are also visible in Fig.~\ref{ablationvis}. In contrast, enabling covariance deformation leads to non-convergent training, supporting our design choice to keep it fixed for stable Gaussian grouping.\\
\textbf{Scale-aware grouping and multiscale training.}
Table~\ref{tab:ablation_sg_mt_psnr} shows that scale-aware grouping improves scalability but remains insufficient on its own, as rendering quality still degrades at low resolutions without multi-scale training. Combining scale-aware grouping with multi-scale training yields the most consistent reconstructions across resolutions, indicating that grouping structures the representation across scales and multi-scale training supplies the supervision needed for robust generalization.
\section{Conclusion}
We proposed D2GV-AR, a deformable 2D Gaussian video representation that supports arbitrary-scale rendering and any-ratio progressive coding from a single model. By combining GoP-level deformation modeling with neural ODE, D2GV-AR enables efficient decoding while matching or surpassing INR-based methods in visual quality. We further established a Nyquist sampling-like rule for grouping primitives to render any resolutions, and formulated progressive pruning as a D-optimal subset selection problem with an efficient score that trades off footprint against overlap redundancy. Extensive experiments demonstrate reliable rate--quality controllability, high rendering throughput, and generalization to downstream tasks.

\section*{Impact Statements}
This paper presents work whose goal is to advance the field of Machine Learning. There are many potential societal consequences of our work, none which we feel must be specifically highlighted here.
\nocite{langley00}

\bibliography{example_paper}
\bibliographystyle{icml2025}

\newpage
\appendix
\onecolumn
\section*{Appendix}
\section{Motivation Recap}
As summarized in Table~\ref{tab:capability}, existing video INRs typically excel in only a subset of capabilities: they offer compact representations but are often tied to a single resolution after training, while multi-resolution variants usually render at only a few preset branches. Moreover, progressive bitrate adaptation is not naturally supported: common INR pruning rules are heuristic and frequently require post-pruning finetuning to recover quality, which is incompatible with on-the-fly switching in video-on-demand and streaming.

We address these limitations by adopting an explicit 2D Gaussian representation. Its parametric primitives enable fast rasterization and provide direct control knobs for scalable rendering. Crucially, we ground these controls in theory: a Nyquist-guided criterion yields content-adaptive, scale-dependent Gaussian grouping for arbitrary target resolutions, and a D-optimal subset selection formulation provides a lightweight proxy for progressive pruning. Together, this enables a single representation to support multi-scale rendering and continuous rate-quality control without maintaining multiple independent models or performing per-ratio finetuning.

\begin{table}[!ht]
\centering
\caption{Capability comparison across methods. \cmark~indicates supported and \xmark~indicates unsupported.}
\label{tab:capability}
\setlength{\tabcolsep}{6pt}
\renewcommand{\arraystretch}{1.12}

\resizebox{0.35\linewidth}{!}{%
\small
\begin{tabular}{@{} l cc c @{}}
\toprule
\multirow{2}{*}{\textsc{Method}} &
\multicolumn{2}{c}{\textsc{Video Representation}} &
\multirow{2}{*}{\textsc{Progressive}} \\
\cmidrule(lr){2-3}
& \textsc{Single} & \textsc{Multiple} & \textsc{Coding} \\
\midrule
\textsc{NeRV}   & \cmark & \cmark & \xmark \\
\textsc{E-NeRV} & \cmark & \xmark & \xmark \\
\textsc{FFNeRV} & \cmark & \xmark & \xmark \\
\textsc{HNeRV}  & \cmark & \xmark & \xmark \\
\textsc{SNeRV}  & \cmark & \cmark & \xmark \\
\textbf{\textsc{Ours}} & \cmark & \cmark & \cmark \\
\bottomrule
\end{tabular}%
}
\end{table}

\section{Experimental Setup}
\subsection{Hyperparameter}
\label{experimental}
We train D2GV at the GoP level with a two-stage pipeline. In the coarse stage, we optimize a set of 2D Gaussians~\cite{zhang2024gaussianimage} to rasterize each frame without deformation, using standard backpropagation. We use a learning rate of \(1\times 10^{-2}\) and train for 5{,}000 epochs. In the fine stage, we initialize from the coarse Gaussians and apply deformation before rasterization; both the deformation MLP and the 2D Gaussians are optimized end-to-end. We reset the Gaussian learning rate to \(5\times 10^{-3}\). For the MLP, we adopt the exponential learning-rate schedule used in Plenoxels, with initial/final rates \(1.6\times 10^{-4}\) and \(1.6\times 10^{-5}\), step size \(1\times 10^{-2}\), and a maximum of 100k steps. We train the fine stage for 120k iterations. The deformation network is a lightweight MLP with a single fully connected layer of width 156, taking the concatenated positional encodings of the Gaussian center and timestamp as input. We use 10 frequencies for spatial encoding and 6 for temporal encoding. We report two model sizes (small 23k primitives/large 46k primitives); the primitive counts and GoP length are summarized in Table~\ref{tab:seq_settings_all}. Note that the primitive counts can vary across sequences, since they depend on both the chosen GoP length and the video duration. For fair comparison, all INR baselines are configured to have a close number of parameters.

\begin{table}[t]
\centering
\caption{Sequence settings for different model sizes (left: 3.5M/5.8M, right: 1.8M/3.3M).}
\label{tab:seq_settings_all}
\setlength{\tabcolsep}{4pt}
\renewcommand{\arraystretch}{1.10}

\resizebox{0.75\linewidth}{!}{%
\small
\begin{tabular}{@{} l l c c c @{\hskip 14pt} l l c c c @{}}
\toprule
\multicolumn{5}{c}{\textsc{Small (3.5M) / Large (5.8M)}} &
\multicolumn{5}{c}{\textsc{Small (1.8M) / Large (3.3M)}} \\
\cmidrule(lr){1-5}\cmidrule(lr){6-10}
\textsc{Model} & \textsc{Scene} & \textsc{\#Frames} & \textsc{GoP} & \textsc{\#Prims} &
\textsc{Model} & \textsc{Scene} & \textsc{\#Frames} & \textsc{GoP} & \textsc{\#Prims} \\
\midrule
\textsc{Small (3.5M)} & \textsc{Bosphorus} & 150 & 10 & 26000 &
\textsc{Small (1.8M)} & \textsc{bmx}   & 80  & 10 & 26000 \\
                       & \textsc{Beauty}    & 150 & 10 & 26000 &
                       & \textsc{boat}  & 75  & 10 & 26000 \\
                       & \textsc{Bee}       & 150 & 10 & 26000 &
                       & \textsc{train} & 80  & 10 & 26000 \\
                       & \textsc{SetGo}     & 150 & 5  & 11000 &
                       & \textsc{camel} & 90  & 10 & 22000 \\
                       & \textsc{Yatcht}    & 150 & 5  & 11000 &
                       & \textsc{cows}  & 103 & 10 & 17000 \\
                       & \textsc{Jockey}    & 150 & 5  & 11000 &
                       & \textsc{swan}  & 50  & 6  & 26000 \\
                       & \textsc{Shake}     & 80  & 5  & 26000 &
                       &                 &     &    &       \\
\midrule
\textsc{Large (5.8M)} & \textsc{Bosphorus} & 150 & 10 & 46000 &
\textsc{Large (3.3M)} & \textsc{bmx}   & 80  & 10 & 46000 \\
                       & \textsc{Beauty}    & 150 & 10 & 46000 &
                       & \textsc{boat}  & 75  & 10 & 46000 \\
                       & \textsc{Bee}       & 150 & 10 & 46000 &
                       & \textsc{train} & 80  & 10 & 46000 \\
                       & \textsc{SetGo}     & 150 & 5  & 21000 &
                       & \textsc{camel} & 90  & 10 & 43000 \\
                       & \textsc{Yatcht}    & 150 & 5  & 21000 &
                       & \textsc{cows}  & 103 & 10 & 35000 \\
                       & \textsc{Jockey}    & 150 & 5  & 21000 &
                       & \textsc{swan}  & 50  & 6  & 46000 \\
                       & \textsc{Shake}     & 80  & 5  & 46000 &
                       &                 &     &    &       \\
\bottomrule
\end{tabular}%
}
\end{table}

\subsection{Size Calculation}
Let a video of length $L$ frames be partitioned into groups of pictures (GoPs) of length $G$ frames, so that the number of GoPs is $N_{\mathrm{GoP}}=\frac{L}{G}$ (assuming $G$ divides $L$). We parameterize each GoP independently using (i) an MLP with $P_{\mathrm{MLP}}$ learnable parameters and (ii) a set of $N_{\mathrm{prim}}$ primitives (e.g., Gaussian splats), where each primitive carries a small fixed set of per-primitive learnable attributes. In our case, each primitive contributes $2$ parameters for its 2D mean (screen-space center) and two $3$-parameter vectors (e.g., scale and color coefficients), yielding $2+3+3$ parameters per primitive. Therefore, the total number of learnable parameters over the whole video scales linearly with the number of GoPs and can be written as
\[
P_{\mathrm{total}}
\;=\;
N_{\mathrm{GoP}}\Big(P_{\mathrm{MLP}} + N_{\mathrm{prim}}(2+3+3)\Big)
\;=\;
\frac{L}{G}\Big(P_{\mathrm{MLP}} + N_{\mathrm{prim}}\cdot 8\Big),
\]
which makes explicit the trade-off between temporal segmentation (smaller $G$ increases $N_{\mathrm{GoP}}$) and per-GoP model capacity.
\begin{figure}[ht]
    \centering
    \includegraphics[width=0.8\linewidth]{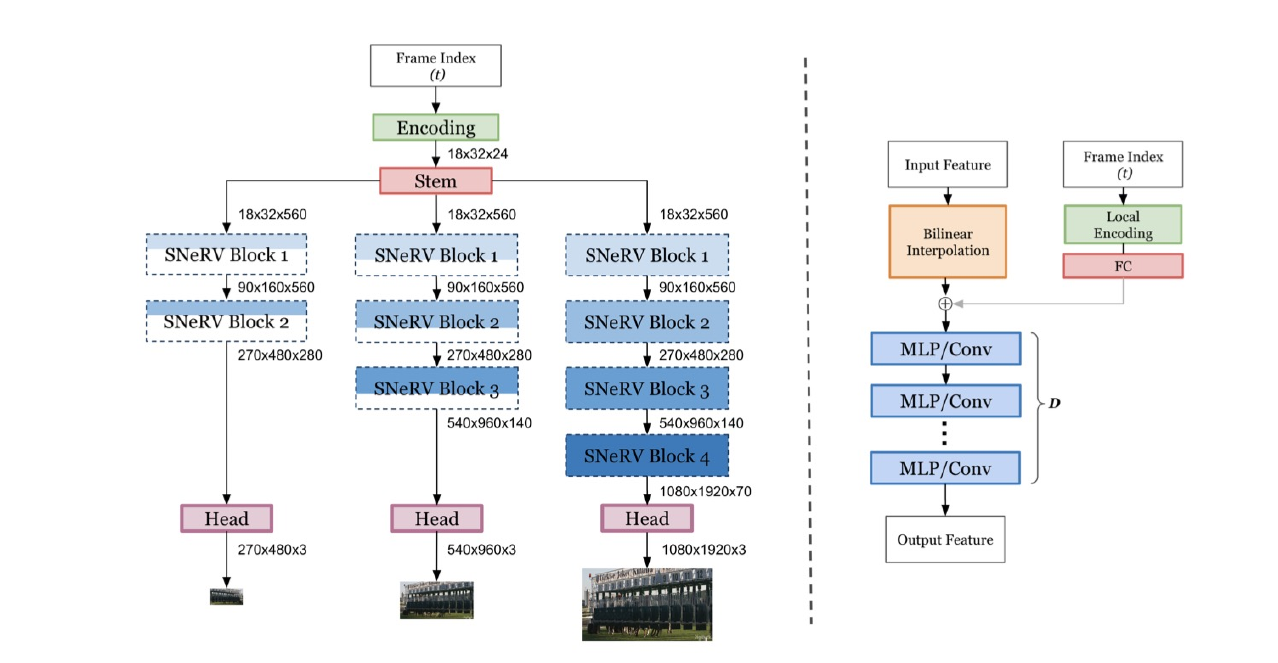}
    \caption{Architecture of SNeRV.}
    \label{snerv archi}
\end{figure}
\subsection{Baseline}
\label{baseline}
For single-resolution comparisons, we use NeRV, E-NeRV, FFNeRV, and HNeRV as INR baselines and reproduce their results using the released code. For the GS-based baseline D-3DGS, we remove perspective projection and annealed smoothing, since these components are tailored to monocular view synthesis rather than 2D video rendering. We also implement a simplified deformation MLP without skip connections and without the Neural ODE formulation.

For multi-resolution baselines, we adopt the official multi-resolution configuration in NeRV (NeRV-MR) \cite{NeRV_GitHub_2021}, which attaches additional output heads after stacking sufficient NeRV blocks to produce multiple resolutions; this can be viewed as a single-trunk design. For SNeRV, since no official code is available, we implement it following the original paper. Specifically, SNeRV uses multiple resolution branches, where higher-resolution branches reuse shallow convolutional kernels from lower-resolution ones to share computation. An illustration of SNeRV architecture is provided in Fig.~\ref{snerv archi}. To our knowledge, these are the latest INRs we could find that support multi-scale rendering.

For the GS-based baseline D-3DGS, we implement it following \cite{yang2024deformable}. Since it is originally designed for multi-view synthesis while our setting focuses on 2D reconstruction, we remove camera projection and the annealed smoothing schedule.

\subsection{Video Compression}
We follow the experimental protocol of GaussianImage~\cite{zhang2024gaussianimage} to compress the canonical Gaussians.
For multi-resolution compression, we only apply quantization-aware attribute fine-tuning so that the quantized representation remains compatible with progressive decoding. We benchmark VP9 scalable video coding using the reference \texttt{libvpx} encoder on raw YUV 4:2:0 sequences at $1280\times720$ and 30\,fps.
To test fine-grained resolution adaptation, we use a five-layer spatial pyramid with non-dyadic scaling factors $\{1/4,\,3/8,\,1/2,\,3/4,\,1\}$, corresponding to $\{320\times180,\,480\times270,\,640\times360,\,960\times540,\,1280\times720\}$.
Rate control is variable bit-rate with CPU speed 4, and cumulative bitrate allocation across layers is set to $\{10\%,\,25\%,\,45\%,\,70\%,\,100\%\}$ of the target bitrate.
We select 500 and 1000\,kbps for comparison and report results on the last three layers ($\times\frac{1}{2}$, $\times\frac{3}{4}$, $\times1$), measuring Overall Y-PSNR against the correspondingly downsampled references.

\section{Ablation Study}
\label{Ablation Study}
We ablate three factors: (i) NeuralODE, (ii) scale-aware grouping and multi-scale training, (iii) the width of the deformation MLP, and (iiii) the coherence weight penalty.

\begin{figure}[!ht]
    \centering
    \includegraphics[width=0.7\linewidth]{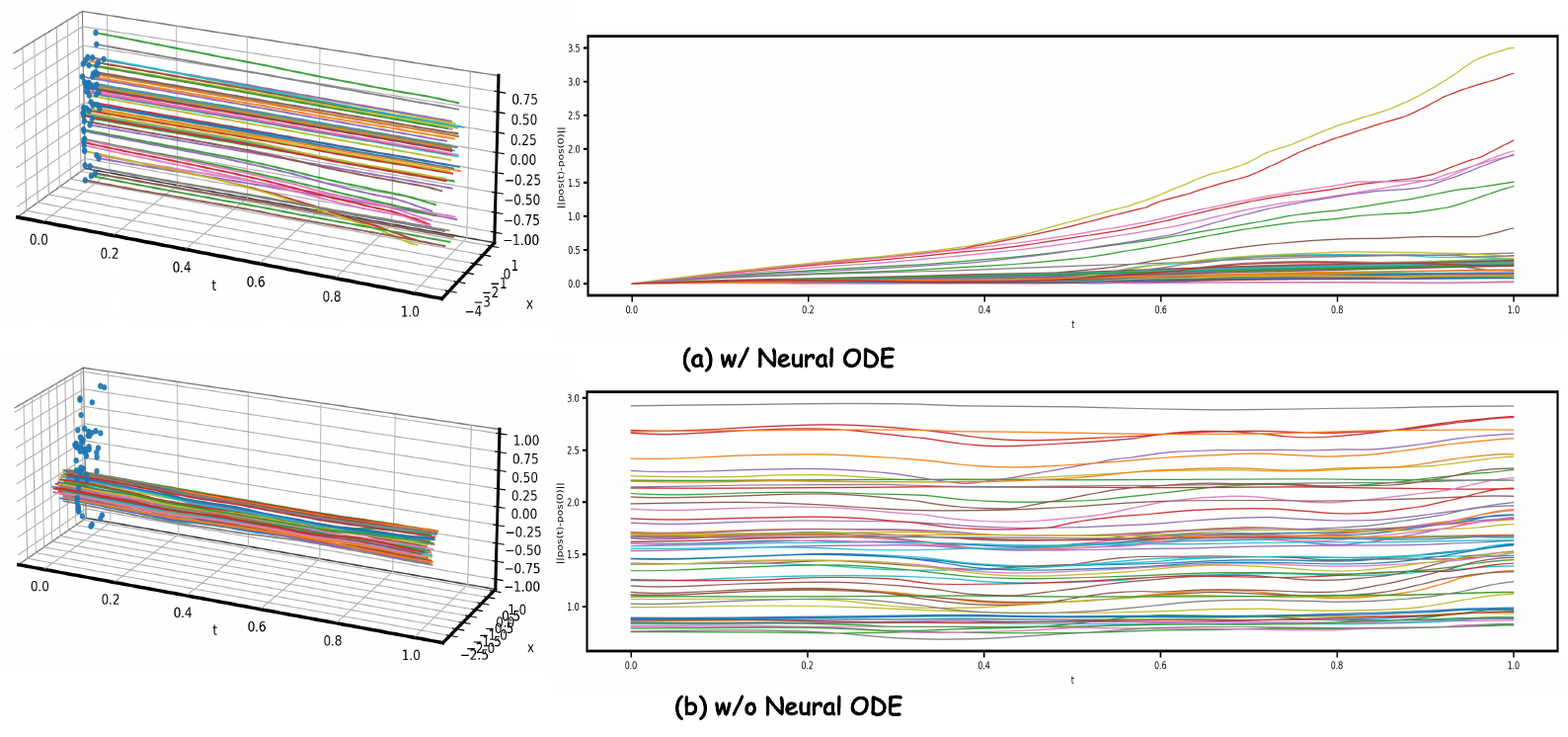}
    \caption{Visualization on 2D Gaussian center deformation.}
    \label{flow}
\end{figure}

\textbf{NeuralODE.} Table~\ref{tab:ablation_neuralode} shows that NeuralODE consistently improves rendering quality, at the cost of lower throughput due to ODE integration overhead. Fig.~\ref{flow} provides a qualitative corroboration: NeuralODE enforces a temporally continuous deformation trajectory, whereas removing it yields timestamp-wise deformations that are less coupled across time.

\begin{figure}[!ht]
    \centering
    \includegraphics[width=0.55\linewidth]{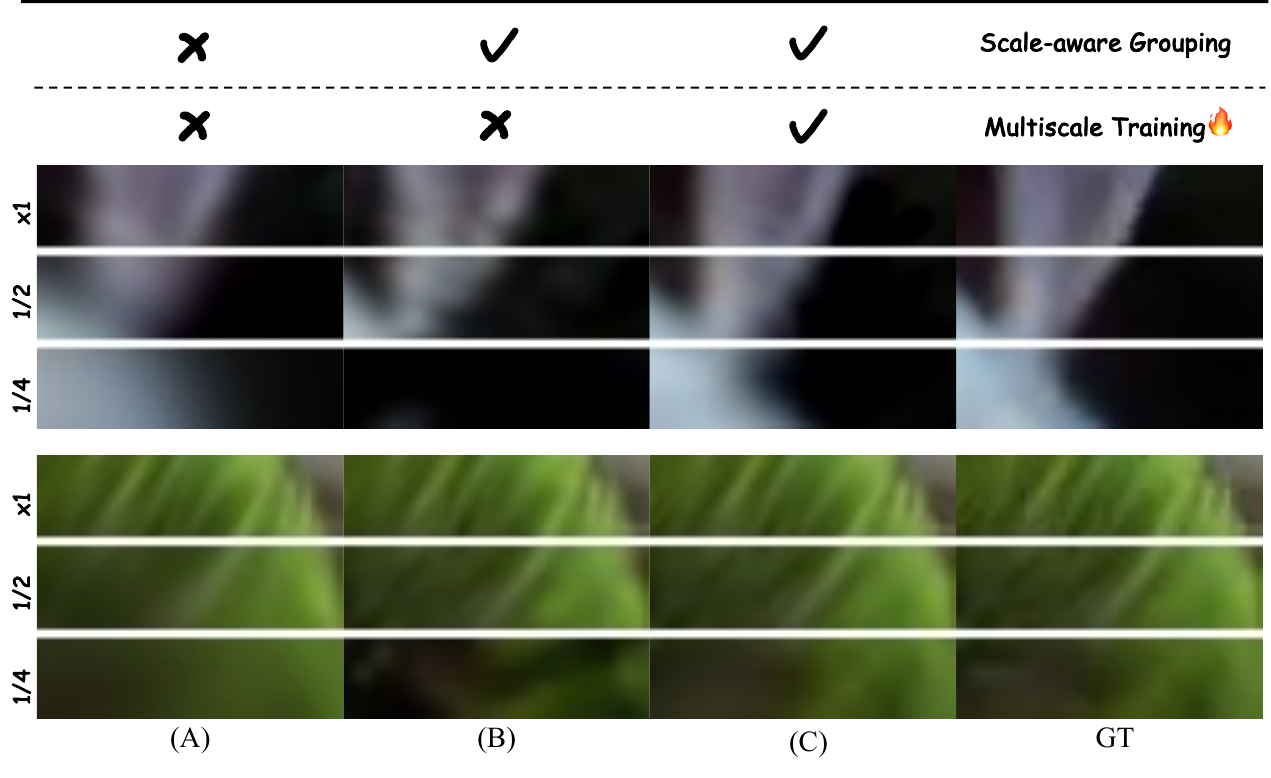}
    \caption{Ablation on multiscale training.}
    \label{mtvis}
\end{figure}

\textbf{Scale-aware grouping and multi-scale training.} Fig.~\ref{mtvis} visualizes their complementary roles. Without grouping, low-resolution rendering exhibits noticeable aliasing. Nyquist-based grouping substantially reduces aliasing, but can leave sparse holes or missing details. Adding multi-scale training mitigates these artifacts and yields the most stable reconstructions across scales.

\textbf{Deformation MLP width.} Table~\ref{tab:mlp_width} indicates a clear quality--efficiency trade-off. Narrower MLPs improve throughput but degrade reconstruction quality, suggesting limited deformation capacity. Increasing width beyond our default brings negligible quality gains while noticeably slowing down decoding. Overall, the default width offers the best balance between fidelity and efficiency.

\textbf{Coherence weight.} 
 in Table \ref{lambda}, we test the sensitivity of coherence weight $\lambda$ and find when it is set as 3 the results are the best. The trend is exptected as small weight could not fully take overlap into consideration while settinhg this factor too large might cause neighboring Gaussians to be discontiuous, leading to holes that degrade the quality. 
 
\begin{table}[ht]
\centering
\caption{Sensitivity of the coherence weight $\lambda_\rho$ in $\mathrm{score}_n=(s_{x,n}s_{y,n})\exp(-\lambda\rho_n)$.
We report BD-rate differences (\%) relative to the area-only baseline; lower is better.}
\label{lambda}
\setlength{\tabcolsep}{6pt}
\renewcommand{\arraystretch}{1.10}
\resizebox{0.25\linewidth}{!}{%
\small
\begin{tabular}{@{}c c@{}}
\toprule
$\lambda_\rho$ & BD-rate vs.\ area-only (\%) $\downarrow$ \\
\midrule
1 & -1.08 \\
2 & -1.45 \\
3 & \textbf{-1.57} \\
4 & -1.35 \\
5 & -0.30 \\
6 & +0.53 \\
7 & +2.14 \\
8 & +3.78 \\
\bottomrule
\end{tabular}%
}
\end{table}

\begin{table}[t]
\centering
\caption{Ablation on Neural ODE. Results are reported on Davis and UVG (Large setting).}
\label{tab:ablation_neuralode}
\setlength{\tabcolsep}{6pt}
\renewcommand{\arraystretch}{1.12}

\resizebox{0.42\linewidth}{!}{%
\small
\begin{tabular}{@{} l l c c c @{}}
\toprule
\textsc{Dataset} & \textsc{Setting} & \textsc{FPS}$\uparrow$ & \textsc{PSNR}$\uparrow$ & \textsc{MS-SSIM}$\uparrow$ \\
\midrule
\multirow{2}{*}{\textsc{Davis}}
& w/o Neural ODE & 308 & 31.09 & 0.9611 \\
& \textbf{Ours}  & 198 & 31.54 & 0.9667 \\
\midrule
\multirow{2}{*}{\textsc{UVG}}
& w/o Neural ODE & 310 & 34.31 & 0.9635 \\
& \textbf{Ours}  & 205 & 35.83 & 0.9727 \\
\bottomrule
\end{tabular}%
}
\end{table}

\begin{table}[!ht]
\centering
\caption{Ablation on MLP width reported as percentage differences (\%) w.r.t.\ width 156 (Ours). Higher is better.}
\label{tab:mlp_width}
\setlength{\tabcolsep}{6pt}
\renewcommand{\arraystretch}{1.12}

\resizebox{0.3\linewidth}{!}{%
\small
\begin{tabular}{@{} l c c c @{}}
\toprule
\textsc{Width} & \textsc{FPS}$\uparrow$ & \textsc{PSNR}$\uparrow$ & \textsc{MS-SSIM}$\uparrow$ \\
\midrule
\textsc{32}  & +30.99 & -6.85 & -0.99 \\
\textsc{64}  & +15.70 & -1.89 & -0.20 \\
\textsc{96}  & +5.58  & -1.34 & -0.10 \\
\textsc{128} & +3.10  & -0.74 & -0.04 \\
\textbf{\textsc{156}} & \textbf{+0.00}  & \textbf{+0.00} & \textbf{+0.00} \\
\textsc{190} & -17.56 & +0.14 & +0.03 \\
\bottomrule
\end{tabular}%
}
\end{table}

\begin{table}[!ht]
\centering
\caption{Bunny multi-resolution comparison (PSNR). Best/second best at each scale are highlighted.}
\label{tab:bunny_mr_psnr_compare}
\setlength{\tabcolsep}{4pt}
\renewcommand{\arraystretch}{1.06}
\resizebox{0.55\linewidth}{!}{%
\small
\begin{tabular}{l ccc ccc ccc}
\toprule
\multirow{2}{*}{\textsc{Model Size}}
& \multicolumn{3}{c}{\textsc{NeRV-MR}}
& \multicolumn{3}{c}{\textsc{SNeRV}}
& \multicolumn{3}{c}{\textsc{Ours-MR}} \\
\cmidrule(lr){2-4}\cmidrule(lr){5-7}\cmidrule(lr){8-10}
& \textbf{$\times$1} & \textbf{$\times$2} & \textbf{$\times$4}
& \textbf{$\times$1} & \textbf{$\times$2} & \textbf{$\times$4}
& \textbf{$\times$1} & \textbf{$\times$2} & \textbf{$\times$4} \\
\midrule
1.5M
& \underline{31.49} & \underline{31.73} & \underline{32.42}
& 30.56 & 29.96 & 30.84
& \textbf{33.22} & \textbf{34.36} & \textbf{33.10} \\
3.0M
& 33.30 & 33.46 & 34.08
& \underline{34.67} & \underline{33.64} & \underline{34.90}
& \textbf{36.83} & \textbf{37.93} & \textbf{36.06} \\
4.5M
& 35.66 & \underline{36.01} & \underline{37.18}
& \underline{35.96} & 35.23 & 36.76
& \textbf{38.23} & \textbf{39.19} & \textbf{37.28} \\
\bottomrule
\end{tabular}%
}
\end{table}

\section{Main Results}
\label{moreresults}
\subsection{More Results on Bunny}
We also compare our method with representative INR baselines on \textit{Bunny} under both single-resolution and multi-resolution settings. In the single-resolution setting, Table~\ref{bunny} reports PSNR on \textit{Bunny} across four model sizes. Our method achieves the best PSNR at every budget, with especially clear gains in the low-capacity regime (e.g., $30.01$~dB at 0.35M versus $28.56$~dB for the strongest baseline). In the multi-resolution setting, Table~\ref{tab:bunny_mr_psnr_compare} reports PSNR at $\times1/\times2/\times4$ for three model sizes. Our method achieves the best PSNR at all three scales across all model sizes, with the strongest gains at larger budgets (e.g., at 4.5M, $38.23/39.19/37.28$ at $\times1/\times2/\times4$). Overall, the results show that our representation provides more consistent quality across resolutions than NeRV-MR and SNeRV.

\begin{table}[t]
\centering
\caption{Single-scale evaluation on \textit{Bunny} dataset.}
\label{bunny}
\setlength{\tabcolsep}{5pt}
\renewcommand{\arraystretch}{1.12}

\resizebox{0.35\linewidth}{!}{%
\small
\begin{tabular}{@{} l c c c c @{}}
\toprule
\textsc{Method} & \textsc{0.35M} & \textsc{0.75M} & \textsc{1.5M} & \textsc{3M} \\
\midrule
\textsc{NeRV}    & 27.50 & 30.36 & 32.34 & 33.46 \\
\textsc{E-NeRV}  & 28.02 & 31.18 & 34.27 & 37.31 \\
\textsc{FFNeRV}  & 28.56 & 31.45 & 34.68 & 37.46 \\
\textsc{HNeRV}   & 27.81 & 32.18 & 35.05 & 37.59 \\
\midrule
\textsc{Ours-SR} & 30.01 & 32.08 & 35.21 & 37.64 \\
\bottomrule
\end{tabular}%
}
\end{table}


\subsection{Per-scene quantitative results}
We report per-scene quantitative results on UVG and DAVIS under both settings.
Multi-resolution results are given in Table~\ref{davisdetailmt}, while single-resolution results are reported in Table~\ref{uvgdetail} and Table~\ref{davisdetail}, respectively.

\subsection{Quality distribution across GoPs}
\begin{figure}[!ht]
    \centering
    \includegraphics[width=0.54\linewidth]{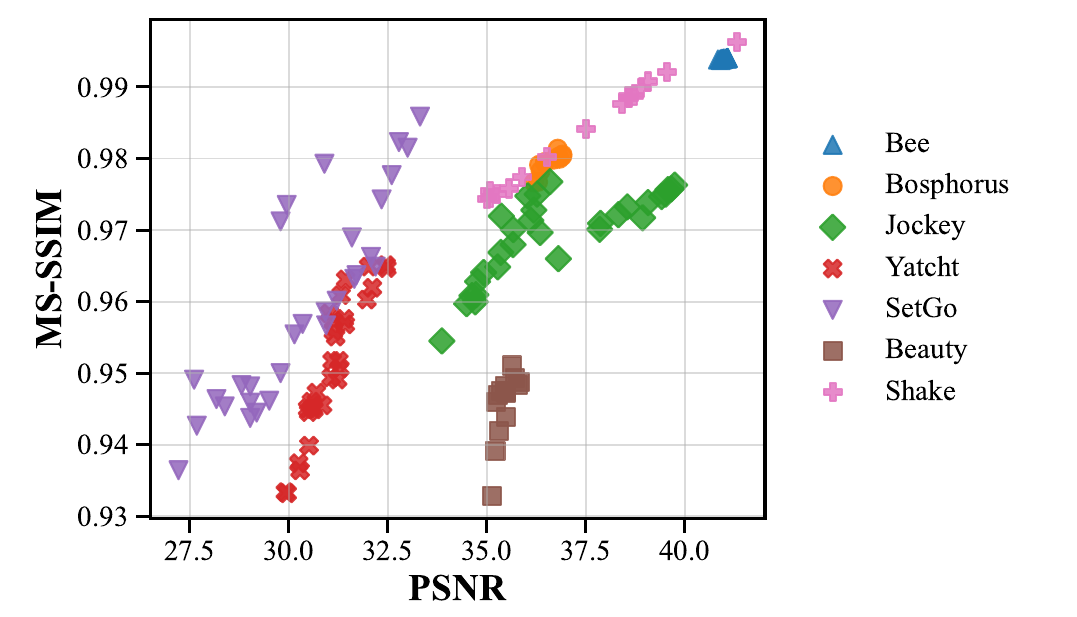}
    \caption{Quality distribution across GoPs on Davis dataset. Model size in large.}
    \label{fig:gop_dist}
\end{figure}

We visualize GoP-level quality on UVG by plotting PSNR against MS-SSIM for seven scenes (Fig.~\ref{fig:gop_dist}). Each point corresponds to one GoP, and points from the same scene form a compact but distinct cluster, indicating that GoP quality is highly content-dependent. Overall, PSNR and MS-SSIM are positively correlated, yet the relationship is not identical across scenes: at similar PSNR, different scenes can exhibit noticeably different MS-SSIM (e.g., \textit{Beauty} vs.\ \textit{Jockey}), suggesting that the perceptual impact of distortion depends on texture and structural complexity.

We also observe that motion intensity and temporal complexity largely govern the within-scene variance. Scenes with more dynamic motion and frequent occlusions tend to show larger GoP-to-GoP fluctuations (e.g., \textit{SetGo} and \textit{Yacht}), spanning a wide range of PSNR and MS-SSIM, which implies that a fixed bitrate may under-serve challenging segments while over-spending on easier ones. In contrast, relatively stable content produces tight clusters with consistently high quality (e.g., \textit{Bee} and \textit{Shake}), indicating more uniform reconstruction across GoPs. These trends motivate GoP-level adaptation: allocating bits according to the GoP difficulty can better match the non-uniform quality distribution observed in practice.

\begin{figure}
    \centering
    \includegraphics[width=0.5\linewidth]{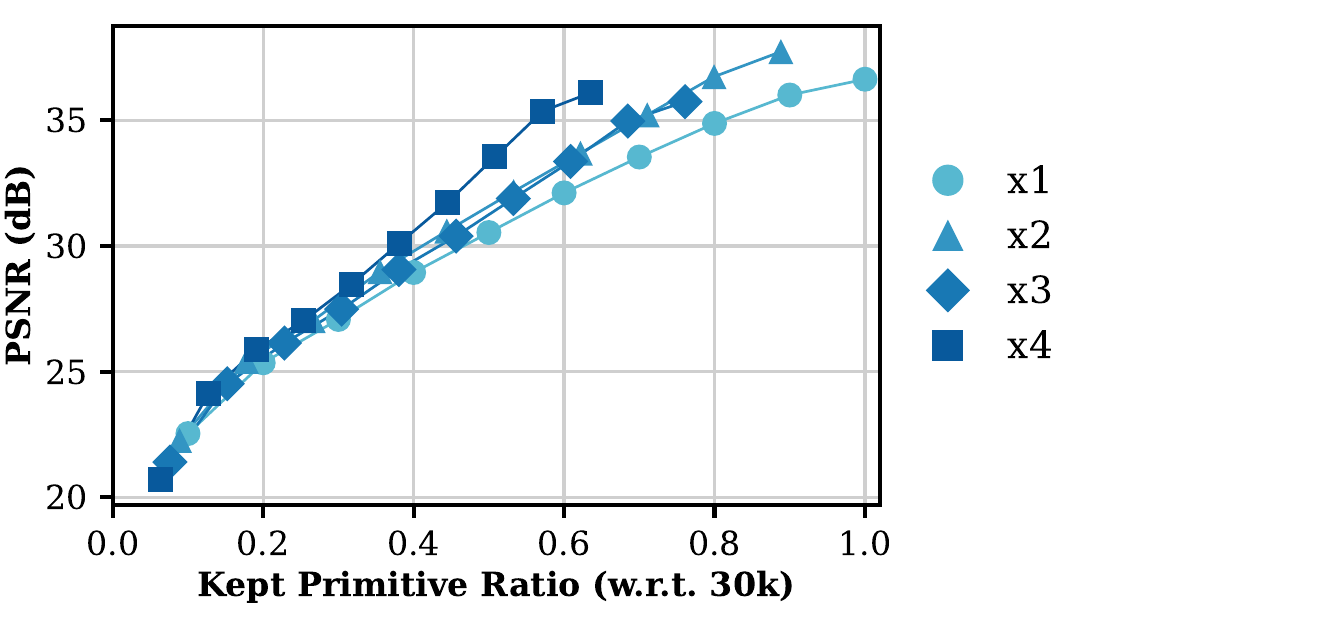}
    \caption{Pruning performance on different resolutions, evaluated on \textit{Bunny} dataset.}
    \label{bunnyprune}
\end{figure}

\begin{figure}[!ht]
    \centering
\includegraphics[width=0.75\linewidth]{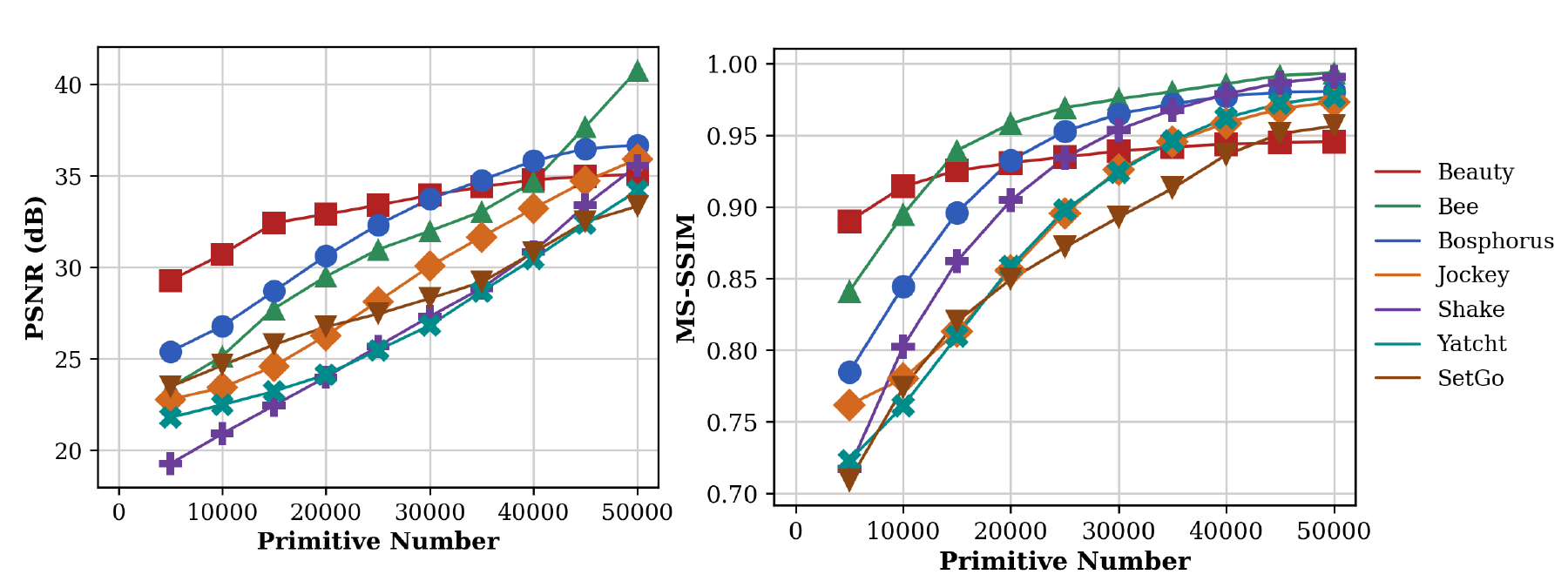}
    \caption{Pruning performance on 720p, evaluated on UVG dataset.}
    \label{uvgprune}
\end{figure}

\begin{figure}[!ht]
    \centering
\includegraphics[width=0.75\linewidth]{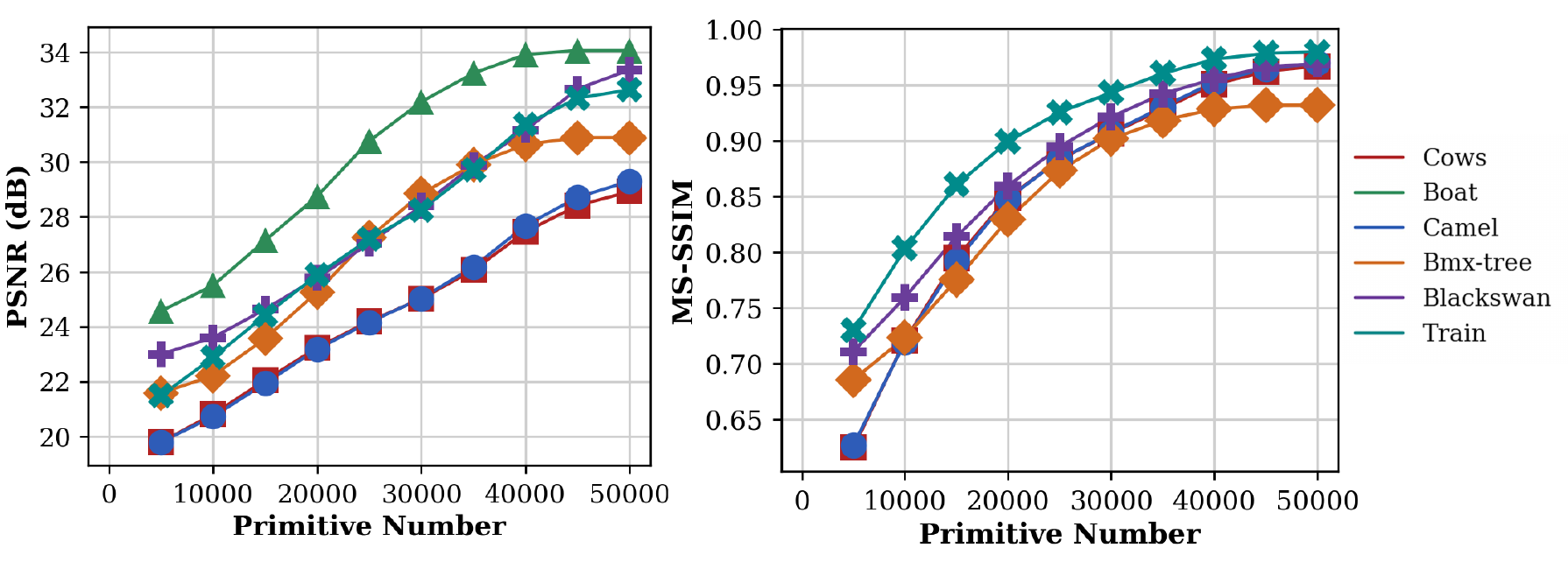}
    \caption{Pruning performance on 720p, evaluated on Davis dataset.}
    \label{davisprune}
\end{figure}

\subsection{Progressive Coding}
\label{pcmore}
\begin{figure*}
    \centering
    \includegraphics[width=\linewidth]{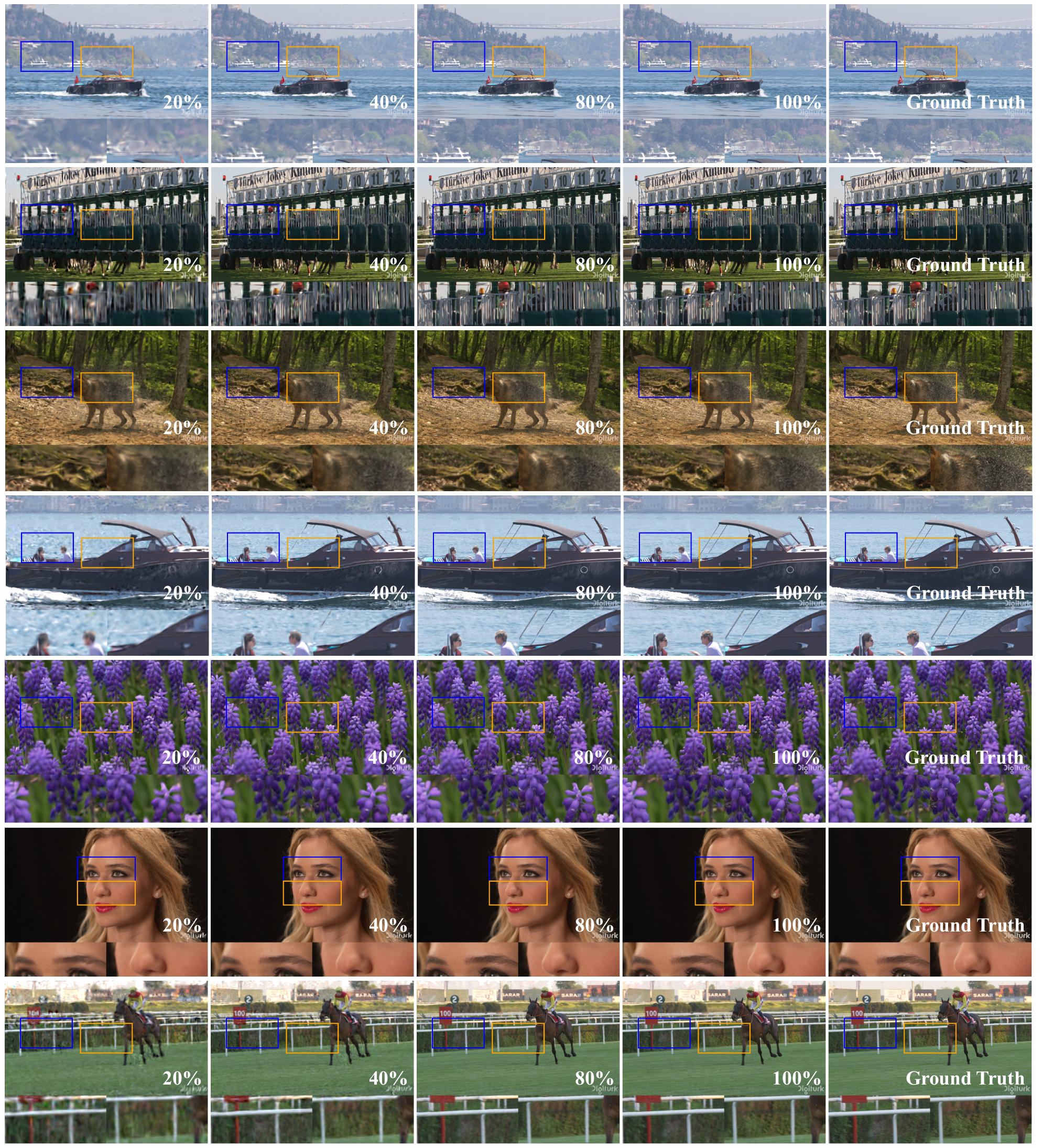}
    \caption{Pruning visualization on UVG dataset.}
    \label{provis1}
\end{figure*}

\begin{figure*}
    \centering
    \includegraphics[width=\linewidth]{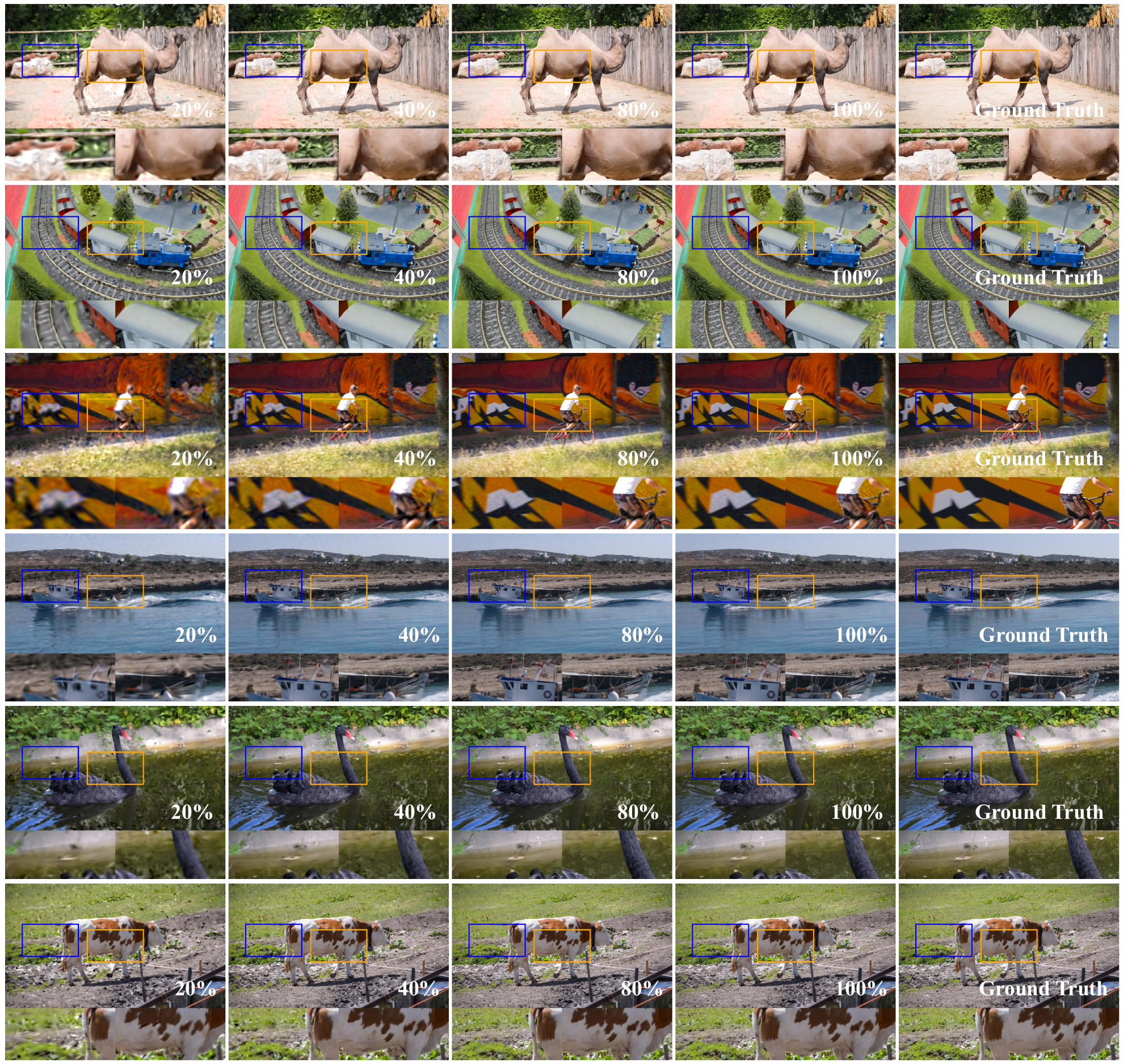}
    \caption{Pruning visualization on Davis dataset.}
    \label{provis2}
\end{figure*}

We summarize the pruning results at different target resolutions in Fig.~\ref{bunnyprune}.
We further aggregate pruning performance over all UVG and DAVIS scenes in Fig.~\ref{uvgprune}, and provide corresponding qualitative comparisons in Fig.~\ref{davisprune}. Qualitative results are provided in Fig. \ref{provis1} and Fig. \ref{provis2}.

\begin{table}[!t]
\centering
\caption{Nomenclature (main symbols).}
\label{tab:nomenclature}
\setlength{\tabcolsep}{6pt}
\renewcommand{\arraystretch}{1.10}
\small
\begin{tabular}{@{} l p{0.78\linewidth} @{}}
\toprule
\textbf{Symbol} & \textbf{Meaning} \\
\midrule
$t,\,t_k$ & Time index / sampled timestamp (often $t_k=k\Delta$). \\
$\Delta$ & Time step between consecutive timestamps. \\
$r$ & Target rendering scale (resolution level). \\
$\mathcal{G}_0^{(r)}$ & Canonical primitive set (Gaussians) available at scale $r$. \\
$N_r$ & Number of primitives at scale $r$, i.e., $N_r=|\mathcal{G}_0^{(r)}|$. \\
$K$ & Decoding/pruning budget (number of selected primitives). \\
$S$ & Selected $K$-subset of primitives, $S\subseteq \mathcal{G}_0^{(r)}$. \\
$\boldsymbol{\mu}$ & 2D Gaussian mean (screen-space center); $\boldsymbol{\mu}_0$ is canonical mean. \\
$\Sigma$ & 2D covariance (shape) of a Gaussian; sometimes stabilized as time-invariant. \\
$\Delta_{\theta}(\cdot)$ & Deformation MLP predicting per-timestamp offsets (discrete-offset baseline). \\
$\mathbf{s}_k$ & Neural ODE latent state at timestamp $t_k$. \\
$\gamma(\cdot)$ & Positional encoding (e.g., Fourier features) for spatial/temporal inputs. \\
$\mathbf{v}_k$ & Discrete velocity induced by offsets, $(\boldsymbol{\mu}_k-\boldsymbol{\mu}_{k-1})/\Delta$. \\
\midrule
$\Omega_t^{(r)}$ & Pixel grid (set of screen pixels) at time $t$ and scale $r$. \\
$\boldsymbol{x}$ & A pixel location on the screen. \\
$\phi(\boldsymbol{x})$ & Induced 2D Gaussian kernel / pixel response at location $\boldsymbol{x}$. \\
$\tau$ & Threshold to define covered pixels (support) from $\phi(\boldsymbol{x})$. \\
$\mathcal{P}(\tau)$ & Covered-pixel set under threshold $\tau$. \\
$\Phi_t$ & Pixel-response matrix at time $t$ (rows: pixels; cols: primitives). \\
$P_t$ & Number of pixels sampled/used at time $t$ for discretization. \\
\midrule
$G$ & Gram matrix of induced kernels (time-aggregated), typically $G=\sum_t \Phi_t^\top \Phi_t$. \\
$G[S]$ & Principal submatrix of $G$ indexed by subset $S$. \\
$\log\det(G[S])$ & D-optimal subset objective for selecting informative, non-redundant primitives. \\
$D$ & Diagonal normalization matrix, $D=\mathrm{diag}(\sqrt{G_{11}},\ldots,\sqrt{G_{N_rN_r}})$. \\
$R$ & Correlation matrix, $R=D^{-1}GD^{-1}$. \\
$R[S]$ & Principal submatrix of $R$ indexed by $S$. \\
$A$ & Off-diagonal overlap matrix, $A=R[S]-I$. \\
$\|\cdot\|_2,\,\|\cdot\|_F$ & Spectral norm and Frobenius norm. \\
$\alpha$ & Bound controlling overlap strength, typically $\|A\|_2\le \alpha<1$. \\
$\lambda_\ell$ & Eigenvalues (e.g., of $A$ in the overlap-term analysis). \\
$\eta$ & Weight balancing coverage vs.\ overlap penalty in the pixel-level surrogate objective. \\
\midrule
$s_x,\,s_y$ & Gaussian scales along principal axes (used in footprint proxy). \\
$\theta$ & In-plane rotation angle in covariance parameterization. \\
$\lambda_\rho$ & Coherence/overlap penalty weight in the lightweight ranking score. \\
$\det(\cdot)$ & Determinant operator. \\
\bottomrule
\end{tabular}
\end{table}

\section{Theoretical Analysis}
In this section, we provide detailed theoretical analyses of Neural ODE modeling and the D-optimal pruning formulation, along with the proof of the Nyquist sampling theorem and the bounded approximation of the overlap term.
\subsection{Velocity-view Analysis: Discrete Offsets vs.\ Neural ODE}
\label{app:vel_view}

We analyze deformation from a \emph{speed} viewpoint under uniformly spaced timestamps.
Let the training times be $t_k = k\Delta$ for $k=0,\dots,T$ with constant step $\Delta$.
We focus on the Gaussian center $\boldsymbol{\mu}$ (the same intuition applies to other attributes).

\paragraph{Per-timestamp prediction yields free discrete speeds.}
A common design predicts an offset independently at each timestamp:
\begin{equation}
\boldsymbol{\mu}_k \;=\; \boldsymbol{\mu}_0 \;+\; \Delta_{\theta}(\boldsymbol{\mu}_0,t_k),
\label{eq:per_time_offset_simple}
\end{equation}
where $\Delta_{\theta}$ is an MLP.
On a uniform grid, the implied \emph{discrete velocity} is simply the finite difference
\begin{equation}
\mathbf{v}_k \;\triangleq\; \frac{\boldsymbol{\mu}_k-\boldsymbol{\mu}_{k-1}}{\Delta}.
\label{eq:discrete_vel}
\end{equation}
Training losses constrain $\boldsymbol{\mu}_k$ at the observed frames, but they do \emph{not} directly constrain the sequence
$\{\mathbf{v}_k\}$.
Thus, even with equal-spaced timestamps, the learned motion can become \emph{variable-speed} and even \emph{jittery}
(large changes in $\mathbf{v}_k$ across $k$), which typically hurts interpolation to unseen timestamps.

In our Neural ODE version, the model does not predict $\boldsymbol{\mu}_k$ independently.
Instead, it predicts a \emph{velocity-like update} at each time and accumulates it over time.
Concretely, we maintain a small latent state $\mathbf{s}_k$ per Gaussian and update it by an MLP:
\begin{equation}
\mathbf{s}_{k+1} \;=\; \mathbf{s}_k \;+\; \Delta \cdot \mathrm{MLP}\big(\gamma(\boldsymbol{\mu}_0),\,\gamma(t_k),\,\mathbf{s}_k\big),
\label{eq:ode_euler_simple}
\end{equation}
and decode the center offset from the state:
\begin{equation}
\boldsymbol{\mu}_k \;=\; \boldsymbol{\mu}_0 \;+\; \mathrm{MLP}_{\text{dec}}(\mathbf{s}_k).
\label{eq:decode_mu_simple}
\end{equation}
This is exactly the discrete-time (Euler) form of a neural ODE.

Because \eqref{eq:ode_euler_simple} adds \emph{incremental} changes, consecutive states $\mathbf{s}_k$ and $\mathbf{s}_{k+1}$ are
coupled by construction.
Therefore, the predicted deformation at $t_{k+1}$ is not free to jump arbitrarily far from $t_k$ unless the MLP outputs a very
large update.
This implicitly discourages sudden speed changes and makes $\{\mathbf{v}_k\}$ in \eqref{eq:discrete_vel} more stable in practice,
which improves generalization to intermediate timestamps: to render an unseen time between $t_k$ and $t_{k+1}$, we can simply
integrate the same update rule with a smaller step size, instead of relying on a separate offset prediction.

\begin{remark}[Constant-speed vs.\ variable-speed]
If the update MLP output is roughly constant over time for a Gaussian, then the accumulated offset grows approximately linearly
with $t$, corresponding to near \emph{constant-speed} motion.
If the output changes with $t$ or the current state, the motion becomes \emph{variable-speed}, but it is still generated by
accumulating small updates, which is typically smoother than independent per-frame offsets.
\end{remark}

\subsection{Problem Formulation for D-optimal Pruning}
\label{app:dopt_from_pixels}

At scale $r$, we decode a $K$-subset $S\subseteq \mathcal{G}_0^{(r)}$ for continuous rate--quality control.
We start from a pixel-intersection view: each primitive induces a screen-space support on the pixel grid, and we want
large total coverage while avoiding counting the same pixels multiple times.
Let $\Omega_t^{(r)}$ be the pixel grid of frame $t$ at scale $r$. For primitive $n$, define its pixel response
\begin{equation}
\label{eq:phi_nt_app}
\phi_{n,t}(\boldsymbol{x})
\triangleq
\exp\!\Big(-\tfrac12(\boldsymbol{x}-\boldsymbol{\mu}_{n,t})^\top \Sigma_{n,t}^{-1}(\boldsymbol{x}-\boldsymbol{\mu}_{n,t})\Big),
\qquad \boldsymbol{x}\in\Omega_t^{(r)} ,
\end{equation}
and the corresponding ``covered-pixel'' set (using a fixed threshold $\tau$)
\begin{equation}
\label{eq:pixel_support_app}
\mathcal{P}_{n,t}(\tau)\triangleq \{\boldsymbol{x}\in\Omega_t^{(r)}:\ \phi_{n,t}(\boldsymbol{x})\ge \tau\}.
\end{equation}
A direct objective is to maximize distinct pixel coverage over time,
\begin{equation}
\label{eq:union_obj_app}
\max_{S:\,|S|=K}\ \sum_{t=1}^T \Big|\ \bigcup_{n\in S}\mathcal{P}_{n,t}(\tau)\ \Big|,
\end{equation}
which explicitly rewards primitives intersecting more pixels and does not double-count pixels explained by multiple
primitives. However,~\eqref{eq:union_obj_app} is discrete and difficult to optimize or rank progressively.

We therefore adopt a smooth surrogate that preserves the same preference: reward total pixel activation while penalizing
redundant co-activation on the same pixels. Using the identity
$\mathbf{1}\{\cup_n \mathcal{P}_{n,t}\}\approx \text{(coverage)}-\text{(overlap)}$,
a standard relaxation is
\begin{equation}
\label{eq:cover_overlap_obj_app}
\max_{S:\,|S|=K}\ 
\sum_{t=1}^T
\Big(
\sum_{n\in S}\!\!\sum_{\boldsymbol{x}\in\Omega_t^{(r)}} \phi_{n,t}(\boldsymbol{x})
\;-\;
\eta \!\!\sum_{\substack{i,j\in S\\ i\neq j}}\ \sum_{\boldsymbol{x}\in\Omega_t^{(r)}} \phi_{i,t}(\boldsymbol{x})\,\phi_{j,t}(\boldsymbol{x})
\Big),
\end{equation}
where the first term is a soft proxy for pixel intersection mass and the second term penalizes multiple primitives
explaining the same pixels.

To expose the kernel subset structure, discretize $\Omega_t^{(r)}$ by pixel locations $\{\boldsymbol{x}_{p,t}\}_{p=1}^{P_t}$
and define $\Phi_t\in\mathbb{R}^{P_t\times N_r}$ with $[\Phi_t]_{p,n}\triangleq \phi_{n,t}(\boldsymbol{x}_{p,t})$.
Then the time-aggregated Gram matrix
\begin{equation}
\label{eq:Gram_time_sum_app}
G \triangleq \sum_{t=1}^T \Phi_t^\top \Phi_t,
\qquad
G_{ij}
=
\sum_{t=1}^T\sum_{p=1}^{P_t}\phi_{i,t}(\boldsymbol{x}_{p,t})\phi_{j,t}(\boldsymbol{x}_{p,t})
\approx
\sum_{t=1}^T\int_{\Omega_t^{(r)}} \phi_{i,t}(\boldsymbol{x})\phi_{j,t}(\boldsymbol{x})\,d\boldsymbol{x}
\end{equation}
summarizes both per-primitive coverage (diagonal) and pairwise redundancy (off-diagonal).
In this form, selecting a subset with large coverage but low redundancy is naturally captured by D-optimal design:
\begin{equation}
\label{eq:dopt_app}
\max_{S:\,|S|=K}\ \log\det\!\big(G[S]\big),
\end{equation}
since $\log\det(G[S])$ increases with the joint span/energy of selected kernel responses and penalizes near-dependence
caused by overlap.

Although $(\boldsymbol{\mu}_{n,t},\Sigma_{n,t})$ are time-varying in general, our implementation stabilizes covariances
to make footprints comparable across frames,
\begin{equation}
\label{eq:Sigma_fixed_over_t_app}
\Sigma_{n,t}=\Sigma_n,\qquad \forall\,t,n.
\end{equation}
With the factorization $G=DRD$, we obtain
\begin{equation}
\label{eq:logdet_decomp_app2}
\log\det\!\big(G[S]\big)=\sum_{n\in S}\log G_{nn}+\log\det\!\big(R[S]\big),
\end{equation}
where the first term describes footprint coverage while $\log\det(R[S])$ is a coupled term describing overall overlap.
Since the second term cannot be assigned to individual primitives, we approximate it via the additive quadratic surrogate
in Lemma~\ref{lem:overlap_surrogate}, which yields the lightweight ranking rule used for progressive transmission.

\subsection{Proof of Nyquist sampling theorem}
\label{Nyquist}
\begin{proof}[Proof of Theorem~\ref{thm:nyquist_gaussian}]
Suppose that $\varepsilon\in(0,1)$. For the Gaussian kernel
$g(x)=\exp\!\big(-x^2/(2\sigma^2)\big)$, its Fourier transform is also a Gaussian:
\[
\hat g(\omega)=C\,\exp\!\Big(-\tfrac{1}{2}\sigma^2\omega^2\Big),
\]
for some constant $C>0$. Hence the magnitude of $\hat g(\omega)$ satisfies
\[
\frac{|\hat g(\omega)|}{|\hat g(0)|}
=\exp\!\Big(-\tfrac{1}{2}\sigma^2\omega^2\Big).
\]
On a grid with spacing $\Delta$, half of the Nyquist frequency is $\omega_N=\pi/\Delta$. Requiring
$|\hat g(\omega_N)|\le \varepsilon\,|\hat g(0)|$ is equivalent to 
\[
\exp\!\Big(-\tfrac{1}{2}\sigma^2\omega_N^2\Big)\le \varepsilon.
\]
Taking logarithms on both sides and rearranging yield
\[
\tfrac{1}{2}\sigma^2\omega_N^2 \ge \ln(1/\varepsilon)
\quad\Longrightarrow\quad
\sigma \ge \frac{\sqrt{2\ln(1/\varepsilon)}}{\omega_N}.
\]
Substituting $\omega_N=\pi/\Delta$ gives
\[
\sigma \ge \frac{\sqrt{2\ln(1/\varepsilon)}}{\pi}\,\Delta
\;\triangleq\;
\beta(\varepsilon)\,\Delta,
\]
which completes the proof.
\end{proof}

\subsection{Proof of Lemma~\ref{lem:overlap_surrogate}.}
\label{proof53}
Fix any subset $S$ with $|S|=K$ and denote
\[
R_S \triangleq R[S]\in\mathbb{R}^{K\times K},\qquad
A \triangleq A_S = R_S - I_K .
\]
Since $R$ is a correlation matrix, $R_S$ is symmetric with $[R_S]_{ii}=1$, hence $A$ is symmetric and
$\mathrm{diag}(A)=\mathbf{0}$.

\smallskip
\noindent\textbf{On the assumption $\|A\|_2\le \alpha<1$.}
The matrix $A=R_S-I_K$ collects the off-diagonal correlations within $S$, so $\|A\|_2$ quantifies the overall strength
of mutual overlap among selected primitives. The condition $\|A\|_2<1$ ensures $R_S=I_K+A\succ 0$ and guarantees the
convergence of the matrix Taylor series for $\log(I_K+A)$.

Moreover, the remainder term in Lemma~\ref{lem:overlap_surrogate} is explicitly controlled by $\alpha$:
combining~\eqref{eq:main_decomp_app} and~\eqref{eq:delta_bound_app} yields the relative bound
\[
\frac{|\Delta(S)|}{\frac12\sum_{\substack{i,j\in S\\ i\neq j}}R_{ij}^2}
\le \frac{2\alpha}{3(1-\alpha)} ,
\]
so the quadratic surrogate is accurate whenever $\alpha$ is moderately small (i.e., $R_S$ is close to $I_K$).

Finally, $\alpha$ can be upper bounded by a simple, computable overlap measure:
\[
\|A\|_2 \le \|A\|_{\infty}
= \max_{i\in S}\sum_{\substack{j\in S\\ j\neq i}} |R_{ij}|,
\]
which is small when no primitive has a large total overlap with many others---exactly the regime encouraged by our
coherence-aware selection.

\smallskip
Assume $\|A\|_2\le \alpha<1$.

\smallskip
Let $\{\lambda_\ell\}_{\ell=1}^K$ be the eigenvalues of $A$. Because $A$ is symmetric,
$|\lambda_\ell|\le \|A\|_2\le \alpha<1$, so the eigenvalues of $R_S=I_K+A$ are $\{1+\lambda_\ell\}$ and satisfy
$1+\lambda_\ell\ge 1-\alpha>0$. Thus $R_S\succ 0$ and $\log\det(R_S)$ is well-defined.

\smallskip
Using $\log\det(R_S)=\mathrm{tr}\big(\log(R_S)\big)$ and the convergent matrix Taylor series
\[
\log(I_K+A)=\sum_{t=1}^{\infty}\frac{(-1)^{t+1}}{t}A^t
\qquad (\text{valid since }\|A\|_2<1),
\]
we get
\begin{equation}
\log\det(R_S)
=\sum_{t=1}^{\infty}\frac{(-1)^{t+1}}{t}\,\mathrm{tr}(A^t).
\label{eq:logdet_series_app}
\end{equation}
Because $\mathrm{diag}(A)=0$, we have $\mathrm{tr}(A)=0$. Moreover, since $A$ is symmetric,
\[
\mathrm{tr}(A^2)=\|A\|_F^2.
\]
Separating the $t=2$ term in~\eqref{eq:logdet_series_app}, define the remainder
\[
\Delta(S)\triangleq \sum_{t=3}^{\infty}\frac{(-1)^{t+1}}{t}\,\mathrm{tr}(A^t),
\]
so that
\begin{equation}
\log\det(R_S)= -\frac12\|A\|_F^2 + \Delta(S).
\label{eq:main_decomp_app}
\end{equation}
Finally, since $A=R_S-I_K$ has zero diagonal and off-diagonal entries $A_{ij}=R_{ij}$ for $i\neq j$ (with indices in $S$),
\[
\|A\|_F^2=\sum_{i,j\in S}A_{ij}^2=\sum_{\substack{i,j\in S\\ i\neq j}}R_{ij}^2,
\]
which proves the first identity in Lemma~\ref{lem:overlap_surrogate}.

\smallskip
Let $\{\lambda_\ell\}_{\ell=1}^K$ be the eigenvalues of $A$. Then for $t\ge 3$,
\[
|\mathrm{tr}(A^t)|
=\Big|\sum_{\ell=1}^K \lambda_\ell^t\Big|
\le \sum_{\ell=1}^K |\lambda_\ell|^t
\le \Big(\max_\ell |\lambda_\ell|\Big)^{t-2}\sum_{\ell=1}^K \lambda_\ell^2
\le \|A\|_2^{\,t-2}\,\|A\|_F^2.
\]
Therefore,
\begin{equation}
|\Delta(S)|
\le \sum_{t=3}^{\infty}\frac{1}{t}\,|\mathrm{tr}(A^t)|
\le \sum_{t=3}^{\infty}\frac{1}{t}\,\|A\|_2^{\,t-2}\,\|A\|_F^2
\le \frac{1}{3}\sum_{t=3}^{\infty}\alpha^{t-2}\,\|A\|_F^2
= \frac{\alpha}{3(1-\alpha)}\,\|A\|_F^2.
\label{eq:delta_bound_app}
\end{equation}
Using $\|A\|_F^2=\sum_{i\neq j\in S}R_{ij}^2$ yields the remainder bound in Lemma~\ref{lem:overlap_surrogate},
completing the proof.
\hfill$\square$

\subsection{Closed-form of 2D Gaussian footprint and correlation.}
\label{area}
For a 2D Gaussian kernel
\begin{equation}
\phi_n(\mathbf{x})=\exp\!\Big(-\tfrac12(\mathbf{x}-\boldsymbol{\mu}_n)^\top\Sigma_n^{-1}(\mathbf{x}-\boldsymbol{\mu}_n)\Big),
\end{equation}
its squared $L_2$ norm has the closed-form expression. It is given in the lemma presented below.

\begin{lemma}[Gaussian $L_2$ norm]
\label{lem:gaussian_l2}
If $\Sigma_n\succ 0$, then
\begin{equation}
\|\phi_n\|_2^2 \;=\; \int_{\mathbb{R}^2}\phi_n(\mathbf{x})^2\,d\mathbf{x}
\;=\; \pi\,\sqrt{\det(\Sigma_n)}.
\end{equation}
Under the rotation--scaling parameterization
$\Sigma_n = R(\theta_n)\,\mathrm{diag}(s_{x,n}^2,s_{y,n}^2)\,R(\theta_n)^\top$,
we have $\sqrt{\det(\Sigma_n)}=s_{x,n}s_{y,n}$.
\end{lemma}

\begin{proof}
We use $\phi_n(\mathbf{x})^2=\exp\!\big(-(\mathbf{x}-\boldsymbol{\mu}_n)^\top\Sigma_n^{-1}(\mathbf{x}-\boldsymbol{\mu}_n)\big)$.
Let $\mathbf{y}=\Sigma_n^{-1/2}(\mathbf{x}-\boldsymbol{\mu}_n)$ so that
$d\mathbf{x}=\sqrt{\det(\Sigma_n)}\,d\mathbf{y}$ and the exponent becomes $-\|\mathbf{y}\|_2^2$.
Hence
\[
\int_{\mathbb{R}^2}\phi_n(\mathbf{x})^2\,d\mathbf{x}
=
\sqrt{\det(\Sigma_n)}\int_{\mathbb{R}^2}e^{-\|\mathbf{y}\|_2^2}\,d\mathbf{y}
=
\sqrt{\det(\Sigma_n)}\cdot \pi,
\]
where $\int_{\mathbb{R}^2}e^{-\|\mathbf{y}\|_2^2}\,d\mathbf{y}=\pi$ is standard.
Finally, $\det(\Sigma_n)=\det(\mathrm{diag}(s_{x,n}^2,s_{y,n}^2))=s_{x,n}^2s_{y,n}^2$
since rotations have determinant $1$.
\end{proof}

For $i,j\in\{1,\dots,N_r\}$, recall
\[
G_{ij}\triangleq \langle \phi_i,\phi_j\rangle=\int_{\mathbb{R}^2}\phi_i(\boldsymbol{x})\phi_j(\boldsymbol{x})\,d\boldsymbol{x},
\qquad
R \triangleq D^{-1}GD^{-1},\ \ R_{ij}=\frac{G_{ij}}{\sqrt{G_{ii}G_{jj}}}.
\]
With $\boldsymbol{d}_{ij}\triangleq \boldsymbol{\mu}_i-\boldsymbol{\mu}_j$, we have the closed form
\begin{equation}
\label{eq:Gij_closed_form}
G_{ij}
=
\frac{2\pi}{\sqrt{\det(\Sigma_i^{-1}+\Sigma_j^{-1})}}
\exp\!\Big(
-\tfrac12\,\boldsymbol{d}_{ij}^{\top}\,
\big(\Sigma_i+\Sigma_j\big)^{-1}\,
\boldsymbol{d}_{ij}
\Big),
\end{equation}
and hence
\begin{equation}
\label{eq:Rij_closed_form}
R_{ij}
=
2\,
\frac{\big(\det(\Sigma_i)\det(\Sigma_j)\big)^{1/4}}{\sqrt{\det(\Sigma_i+\Sigma_j)}}
\exp\!\Big(
-\tfrac12\,\boldsymbol{d}_{ij}^{\top}\,
\big(\Sigma_i+\Sigma_j\big)^{-1}\,
\boldsymbol{d}_{ij}
\Big).
\end{equation}
In particular, since $G_{ii}=\|\phi_i\|_2^2=\pi\sqrt{\det(\Sigma_i)}=\pi s_{x,i}s_{y,i}$, the normalization is consistent with
Lemma~\ref{lem:gaussian_l2}.


\newpage

\begin{table}[t]
\centering
\caption{Multi-resolution results on DAVIS (in large size setting, 3.3M). Scales correspond to 720p (x1), 360p (x2), and 180p (x4). MS-SSIM unavailable is denoted as ``-''.}
\label{davisdetailmt}
\setlength{\tabcolsep}{4pt}
\renewcommand{\arraystretch}{1.10}
\resizebox{0.6\linewidth}{!}{%
\scriptsize
\begin{tabular}{l c || cc | cc | cc}
\toprule
\multirow{2}{*}{\textbf{Scene}} & \multirow{2}{*}{\textbf{Scale}}
& \multicolumn{2}{c|}{\textbf{NeRV-MR}}
& \multicolumn{2}{c|}{\textbf{SNeRV}}
& \multicolumn{2}{c}{\textbf{Ours-MR}} \\
\cmidrule(lr){3-4}\cmidrule(lr){5-6}\cmidrule(lr){7-8}
& & PSNR$\uparrow$ & MS-SSIM$\uparrow$
  & PSNR$\uparrow$ & MS-SSIM$\uparrow$
  & PSNR$\uparrow$ & MS-SSIM$\uparrow$ \\
\midrule

\multirow{3}{*}{bmx}
& x1 & 27.35 & 0.8940 & 29.61 & \textbf{0.9391} & \textbf{30.93} & 0.9326 \\
& x2 & 27.51 & 0.9418 & 30.19 & \textbf{0.9723} & \textbf{31.67} & 0.9641 \\
& x4 & 28.00 & 0.9730 & \textbf{31.91} & \textbf{0.9900} & 30.40 & 0.9767 \\
\midrule

\multirow{3}{*}{boat}
& x1 & 31.42 & 0.9504 & 31.46 & 0.9568 & \textbf{34.10} & \textbf{0.9706} \\
& x2 & 32.30 & 0.9762 & 32.34 & 0.9793 & \textbf{35.78} & \textbf{0.9868} \\
& x4 & 34.32 & 0.9906 & \textbf{34.80} & \textbf{0.9918} & 34.24 & 0.9894 \\
\midrule

\multirow{3}{*}{camel}
& x1 & 24.87 & 0.8953 & 26.87 & 0.9353 & \textbf{28.86} & \textbf{0.9413} \\
& x2 & 25.57 & 0.9498 & 27.13 & 0.9688 & \textbf{29.91} & \textbf{0.9706} \\
& x4 & 27.51 & 0.9793 & \textbf{29.61} & \textbf{0.9883} & 27.90 & 0.9758 \\
\midrule

\multirow{3}{*}{cows}
& x1 & 22.85 & 0.8276 & 24.68 & 0.9062 & \textbf{29.68} & \textbf{0.9728} \\
& x2 & 23.72 & 0.9108 & 25.43 & 0.9533 & \textbf{30.83} & \textbf{0.9837} \\
& x4 & 25.79 & 0.9629 & \textbf{28.32} & \textbf{0.9816} & 27.34 & 0.9736 \\
\midrule

\multirow{3}{*}{swan}
& x1 & 28.18 & 0.9148 & 29.68 & 0.9454 & \textbf{33.70} & \textbf{0.9711} \\
& x2 & 28.33 & 0.9553 & 30.11 & 0.9754 & \textbf{34.83} & \textbf{0.9877} \\
& x4 & 29.08 & 0.9808 & 32.05 & \textbf{0.9911} & \textbf{32.11} & 0.9875 \\
\midrule

\multirow{3}{*}{train}
& x1 & 26.66 & 0.9069 & 27.72 & 0.9266 & \textbf{32.79} & \textbf{0.9807} \\
& x2 & 26.75 & 0.9520 & 28.06 & 0.9660 & \textbf{33.54} & \textbf{0.9906} \\
& x4 & 27.70 & 0.9784 & 29.98 & 0.9877 & \textbf{30.05} & \textbf{0.9884} \\
\midrule

\multirow{3}{*}{\textbf{avg.}}
& x1 & 27.08 & 0.9082 & 28.34 & 0.9349 & \textbf{31.68} & \textbf{0.9615} \\
& x2 & 27.73 & 0.9531 & 28.88 & 0.9692 & \textbf{32.76} & \textbf{0.9806} \\
& x4 & 29.17 & 0.9799 & \textbf{31.11} & \textbf{0.9884} & 30.34 & 0.9819 \\
\bottomrule
\end{tabular}%
}
\end{table}

\begin{table*}[t]
\centering
\caption{Video regression results on small size at resolution 1280x720. Best/second best in each row (per metric) are highlighted.}
\label{uvgdetail}
\setlength{\tabcolsep}{4pt}
\renewcommand{\arraystretch}{1.05}

\resizebox{\textwidth}{!}{%
\scriptsize
\begin{tabular}{ll*{6}{cc}}
\toprule
\textbf{Dataset} & \textbf{Sequence}
& \multicolumn{2}{c}{\textbf{NeRV}}
& \multicolumn{2}{c}{\textbf{E-NeRV}}
& \multicolumn{2}{c}{\textbf{FFNeRV}}
& \multicolumn{2}{c}{\textbf{HNeRV}}
& \multicolumn{2}{c}{\textbf{D-3DGS}}
& \multicolumn{2}{c}{\textbf{Ours-SR}} \\
\cmidrule(lr){3-4}\cmidrule(lr){5-6}\cmidrule(lr){7-8}\cmidrule(lr){9-10}\cmidrule(lr){11-12}\cmidrule(lr){13-14}
& & PSNR & MS-SSIM & PSNR & MS-SSIM & PSNR & MS-SSIM & PSNR & MS-SSIM & PSNR & MS-SSIM & PSNR & MS-SSIM \\
\midrule

\multirow{7}{*}{\textbf{Davis}} & Bmx-tree
& 26.49 & 0.8702 & 27.64 & 0.9009 & \underline{29.06} & \textbf{0.9290} & \textbf{29.61} & 0.9055 & 25.89 & 0.8501 & 28.97 & \underline{0.9125} \\
& Boat
& 30.78 & 0.9388 & 31.36 & 0.9508 & 32.51 & \underline{0.9617} & \textbf{33.85} & 0.9549 & 27.89 & 0.9045 & \underline{33.29} & \textbf{0.9632} \\
& Camel
& 24.14 & 0.8665 & 25.03 & 0.8989 & 26.13 & \underline{0.9153} & \underline{27.22} & 0.9085 & 22.86 & 0.8189 & \textbf{28.94} & \textbf{0.9555} \\
& Cows
& 22.32 & 0.7825 & 23.50 & 0.8513 & 23.90 & \underline{0.8646} & \underline{24.78} & 0.8386 & 21.54 & 0.7689 & \textbf{26.38} & \textbf{0.9319} \\
& Blackswan
& 27.09 & 0.8835 & 28.85 & 0.9264 & 29.96 & \underline{0.9468} & \textbf{31.84} & 0.9433 & 25.95 & 0.8592 & \underline{31.51} & \textbf{0.9533} \\
& Train
& 25.67 & 0.8781 & 26.35 & 0.8983 & 27.62 & \underline{0.9280} & \underline{28.54} & 0.9185 & 23.12 & 0.8293 & \textbf{29.80} & \textbf{0.9579} \\
& \textbf{Average}
& 26.08 & 0.8699 & 27.12 & 0.9044 & 28.20 & \underline{0.9242} & \underline{29.31} & 0.9115 & 24.54 & 0.8385 & \textbf{29.82} & \textbf{0.9457} \\
\midrule

\multirow{8}{*}{\textbf{UVG}} & Beauty
& 34.91 & 0.9444
& \underline{35.68} & \underline{0.9492}
& 35.60 & \textbf{0.9533}
& \textbf{35.90} & 0.9489
& 31.21 & 0.9187
& 35.38 & 0.9468 \\

& Setgo
& 26.23 & 0.9252
& 28.31 & 0.9541
& 28.64 & \underline{0.9579}
& \underline{30.08} & 0.9484
& 24.23 & 0.9043
& \textbf{30.58} & \textbf{0.9748} \\

& Bee
& 39.60 & 0.9922
& 39.67 & 0.9924
& \underline{41.03} & \underline{0.9941}
& \textbf{41.39} & \textbf{0.9942}
& 23.67 & 0.8864
& 39.88 & 0.9931 \\

& Jockey
& 34.04 & 0.9497
& \underline{36.04} & \underline{0.9678}
& \textbf{36.55} & \textbf{0.9703}
& 35.72 & 0.9464
& 26.48 & 0.8751
& 35.48 & 0.9572 \\

& Shake
& 34.70 & 0.9660
& 27.70 & 0.9854
& 36.55 & \underline{0.9819}
& \textbf{38.36} & \textbf{0.9863}
& 22.51 & 0.8157
& \underline{36.78} & 0.9756 \\

& Bosphorus
& 33.56 & 0.9525
& 36.07 & 0.9727
& \textbf{36.72} & \textbf{0.9773}
& 36.47 & 0.9703
& 29.89 & 0.9102
& \underline{36.60} & \underline{0.9745} \\

& Yacht
& 28.27 & 0.9105
& 30.90 & \underline{0.9508}
& \underline{31.10} & \textbf{0.9558}
& \textbf{31.23} & 0.9301
& 25.89 & 0.8973
& 30.51 & 0.9424 \\

& \textbf{Average}
& 33.04 & 0.9486
& 33.48 & \underline{0.9675}
& \underline{35.17} & \textbf{0.9701}
& \textbf{35.59} & 0.9607
& 26.27 & 0.8868
& 35.03 & 0.9663 \\
\bottomrule
\end{tabular}%
}
\end{table*}

\begin{table*}[t]
\centering
\caption{Video regression results on \textbf{large size} at resolution 1280x720. Best/second best in each row (per metric) are highlighted.}
\label{davisdetail}
\setlength{\tabcolsep}{4pt}
\renewcommand{\arraystretch}{1.05}

\resizebox{\textwidth}{!}{%
\scriptsize
\begin{tabular}{ll*{6}{cc}}
\toprule
\textbf{Dataset} & \textbf{Sequence}
& \multicolumn{2}{c}{\textbf{NeRV}}
& \multicolumn{2}{c}{\textbf{E-NeRV}}
& \multicolumn{2}{c}{\textbf{FFNeRV}}
& \multicolumn{2}{c}{\textbf{HNeRV}}
& \multicolumn{2}{c}{\textbf{D-3DGS}}
& \multicolumn{2}{c}{\textbf{Ours-SR}} \\
\cmidrule(lr){3-4}\cmidrule(lr){5-6}\cmidrule(lr){7-8}\cmidrule(lr){9-10}\cmidrule(lr){11-12}\cmidrule(lr){13-14}
& & PSNR & MS-SSIM & PSNR & MS-SSIM & PSNR & MS-SSIM & PSNR & MS-SSIM & PSNR & MS-SSIM & PSNR & MS-SSIM \\
\midrule

\multirow{7}{*}{\textbf{Davis}} & Bmx-tree
& 27.04 & 0.8897 & 29.27 & 0.9325 & \underline{30.63} & \textbf{0.9509} & \textbf{32.19} & \underline{0.9488} & 26.80 & 0.8852 & 30.10 & 0.9341 \\
& Boat
& 31.69 & 0.9482 & 32.94 & 0.9645 & 34.39 & \underline{0.9747} & \textbf{36.03} & 0.9737 & 32.94 & 0.9645 & \underline{34.87} & \textbf{0.9749} \\
& Camel
& 24.85 & 0.8860 & 26.58 & 0.9293 & 27.43 & \underline{0.9410} & \underline{29.14} & 0.9373 & 23.38 & 0.8549 & \textbf{30.97} & \textbf{0.9724} \\
& Cows
& 22.74 & 0.8099 & 22.65 & 0.8102 & 25.35 & \underline{0.9086} & \underline{26.32} & 0.8902 & 22.45 & 0.8099 & \textbf{28.81} & \textbf{0.9657} \\
& Blackswan
& 28.12 & 0.9061 & 30.51 & 0.9502 & 31.59 & 0.9659 & \textbf{34.90} & \textbf{0.9734} & 26.81 & 0.8894 & \underline{33.56} & \underline{0.9724} \\
& Train
& 26.47 & 0.9012 & 29.38 & \underline{0.9524} & 28.97 & 0.9495 & \underline{30.60} & 0.9518 & 23.80 & 0.8452 & \textbf{32.89} & \textbf{0.9811} \\
& \textbf{Average}
& 26.82 & 0.8902 & 28.55 & 0.9232 & 29.73 & \underline{0.9484} & \underline{31.53} & 0.9459 & 26.48 & 0.8808 & \textbf{31.87} & \textbf{0.9668} \\
\midrule

\multirow{8}{*}{\textbf{UVG}} & Beauty
& 35.24 & 0.9446 & \underline{35.82} & 0.9508 & 35.68 & \underline{0.9544} & \textbf{36.33} & \textbf{0.9549} & 32.64 & 0.9268 & 35.58 & 0.9498 \\
& Setgo
& 27.00 & 0.9324 & 30.30 & 0.9689 & 29.68 & 0.9658 & \textbf{32.81} & \underline{0.9730} & 25.41 & 0.9291 & \underline{32.27} & \textbf{0.9821} \\
& Bee
& 39.88 & 0.9924 & 39.06 & \underline{0.9937} & \underline{41.47} & \textbf{0.9944} & \textbf{41.54} & \textbf{0.9944} & 29.21 & 0.9488 & 41.05 & 0.9940 \\
& Jockey
& 34.07 & 0.9509 & 36.09 & \underline{0.9710} & 36.48 & 0.9688 & \textbf{38.74} & \textbf{0.9778} & 27.01 & 0.8831 & \underline{36.74} & 0.9693 \\
& Shake
& 34.98 & 0.9667 & 36.53 & 0.9790 & \textbf{37.77} & \textbf{0.9867} & 37.20 & 0.9813 & 27.73 & 0.8635 & \underline{37.44} & \underline{0.9838} \\
& Bosphorus
& 33.95 & 0.9567 & 36.01 & 0.9760 & \underline{37.23} & 0.9772 & \textbf{38.52} & \textbf{0.9820} & 31.01 & 0.9355 & 36.68 & \underline{0.9793} \\
& Yacht
& 28.67 & 0.9212 & 31.34 & \underline{0.9590} & \underline{31.79} & 0.9569 & \textbf{35.15} & \textbf{0.9720} & 26.78 & 0.9001 & 31.07 & 0.9511 \\
& \textbf{Average}
& 33.40 & 0.9521 & 35.02 & 0.9712 & 35.73 & 0.9720 & \textbf{37.18} & \textbf{0.9765} & 28.54 & 0.9124 & \underline{35.83} & \underline{0.9728}\\
\bottomrule
\end{tabular}%
}
\end{table*}



\end{document}